\documentclass[lettersize,journal]{IEEEtran}
\usepackage{cite}
\usepackage{graphicx}
\usepackage{subcaption}
\usepackage{algorithmic}

\usepackage{textcomp}
\usepackage{xcolor}
\usepackage{amsmath,amssymb,amsfonts}
\usepackage{amsthm} 

\usepackage{algorithm}
\usepackage{booktabs}
\usepackage{placeins}
\usepackage{tabularx}
\usepackage{threeparttable}
\newtheorem{definition}{Definition}
\usepackage{textcomp}
\usepackage{multirow}
\usepackage{pifont}     
\usepackage{amssymb}    
\usepackage{siunitx}    
\usepackage{makecell}
\usepackage{orcidlink}
\usepackage{booktabs}
\usepackage{float}
\usepackage{caption}
\usepackage{booktabs}
\usepackage{array}
\usepackage{tabularx}
\usepackage{lipsum} 
\usepackage{dblfloatfix} 
\usepackage{url}
\usepackage{longtable}

\newcommand{\stepcomment}[1]{
    \STATE /* --- #1 \hfill --- */
}

\begin{document}
\title{GBSK: Skeleton Clustering via Granular-ball Computing and Multi-Sampling for Large-Scale Data} 
\author{
Yewang Chen$^{\orcidlink{0000-0001-9691-0807}}$, \IEEEmembership{Senior Member, IEEE},
Junfeng Li$^{\orcidlink{0009-0000-1456-8738}}$,
Shuyin Xia$^{\orcidlink{0000-0001-5993-9563}}$,\IEEEmembership{Member, IEEE},
Qinghong Lai$^{\orcidlink{0009-0003-0712-9229}}$,
Xinbo Gao$^{\orcidlink{0000-0002-7985-0037}}$,\IEEEmembership{Fellow, IEEE},
Guoyin Wang$^{\orcidlink{0000-0002-8521-5232}}$,\IEEEmembership{Senior Member, IEEE},
Dongdong Cheng$^{\orcidlink{0000-0003-3500-5461}}$,
Yi Liu$^{\orcidlink{0009-0008-9668-7076}}$,
Yi Wang$^{\orcidlink{0000-0002-9013-7232}}$
\thanks{Yewang Chen is with the College of Computer Science and Technology, Huaqiao University, Xiamen, Fujian, China, and the Fujian Provincial Key Laboratory of Computer Vision and Machine Learning, Huaqiao University, Xiamen, Fujian, China (e-mail: ywchen@hqu.edu.cn).}
\thanks{Qinghong Lai, and Junfeng Li are with the College of Computer Science and Technology, Huaqiao University, Xiamen, Fujian, China (e-mail: keenorlai@stu.hqu.edu.cn, marveenlee@stu.hqu.edu.cn).}
\thanks{Shuyin Xia is with the Key Laboratory of Cyberspace Big Data Intelligent Security, Ministry of Education; Chongqing Key Laboratory of Computational Intelligence, Chongqing University of Posts and Telecommunications; and Chongqing Key Laboratory of Big Data Intelligent Computing, Chongqing University of Posts and Telecommunications, Chongqing, China (e-mail: xiasy@cqupt.edu.cn).}
\thanks{Xinbo Gao is with the Chongqing Key Laboratory of Computational Intelligence, Chongqing University of Posts and Telecommunications, Chongqing, China (e-mail: gaoxb@cqupt.edu.cn).}
\thanks{Guoyin Wang is with the College of Computer and Information  Science, Chongqing Normal University, Chongqing, China (e-mail: wanggy@cqnu.edu.cn).}
\thanks{Dongdong Cheng is with the College of Big Data and Intelligent Engineering, Yangtze Normal University, Chongqing, China (e-mail: cdd@yznu.edu.cn).}
\thanks{Yi Liu, and Yi Wang are with Chongqing Ant Consumer Finance Co., Ltd., Ant Group, Chongqing, China (e-mail: larry.liuy@myxiaojin.cn, haonan.wy@myxiaojin.cn).}
\thanks{This work has been submitted to IEEE TPAMI for possible publication.}
}

\markboth{IEEE Transactions on Pattern Analysis and Machine Intelligence}%
{ \MakeLowercase{\textit{et al.}}: GBSK: Skeleton Clustering via Granular-ball Computing and Multi-Sampling for Large-Scale Data} 

\maketitle
\begin{abstract}
To effectively handle clustering task for large-scale datasets, we propose a novel scalable skeleton clustering algorithm, namely GBSK, which leverages the granular-ball technique to capture the underlying structure of data. By multi-sampling the dataset and constructing multi-grained granular-balls, GBSK progressively uncovers a statistical ``skeleton'' — a spatial abstraction that approximates the essential structure and distribution of the original data. This strategy enables GBSK to dramatically reduce computational overhead while maintaining high clustering accuracy. In addition, we introduce an adaptive version, AGBSK, with simplified parameter settings to enhance usability and facilitate deployment in real-world scenarios. Extensive experiments conducted on standard computing hardware demonstrate that GBSK achieves high efficiency and strong clustering performance on large-scale datasets, including one with up to 100 million instances across 256 dimensions. Our implementation and experimental results are available at: \url{https://github.com/XFastDataLab/GBSK/}.
\end{abstract}

\begin{IEEEkeywords}
granular-ball, skeleton clustering, multiple sampling, density peaks.  
\end{IEEEkeywords}

\section{Introduction}
Clustering provides a method for grouping unsupervised data into distinct categories. It involves identifying patterns of similarity or relevance within the dataset, thereby grouping data points into subsets that share similar characteristics \cite{Li2023FuzzyEC, Zhou2020SelfPacedCE,Yang2022RobustMC}. Clustering helps uncover hidden patterns within data, fostering enhanced understanding and utilization\cite{Xu2018ScaledSR,Zhang2020GeneralizedLM,10719619,jiao2024end,nie2020detecting}. A host of algorithms have been developed to handle this task, such as k-means\cite{forgy1965cluster}, Meanshift\cite{meanshift1995}, Spectral Clustering\cite{XuelongLi2023spectral}, AP Clustering\cite{APclustering2013}, DBSCAN\cite{ester1996density}, and Density Peaks (DPeak or DPC) clustering \cite{rodriguez2014clustering}. 

However, these algorithms are unsuitable for large-scale data due to their high computational complexity. For example, 
k-means runs in $O(ktn)$ expected time, where $k$ is the number of clusters, $t$ is the number of iterations, and $n$ is the cardinality of the dataset \cite{hartigan1979algorithm}; 
the complexities of DBSCAN and DPeak are both $O(n^2)$ \cite{Chen2021KNN,dpc2024analysis}. 

To overcome this limitation, various techniques have been proposed, including fast k-nearest neighbors (kNN)-based methods \cite{chen2020fast}, parallel technologies \cite{amagata2021fast}, grid-based methods \cite{gan2015dbscan}, granular-ball-based clustering \cite{cheng2023fast}, and sampling-based methods \cite{tobin2021dcf,liao2013sample,ding2023sampling}. Of these, granular-ball-based and sampling-based methods offer promising solutions. 

However, identifying all the granular-balls within a very-large-scale dataset presents a formidable challenge. The sheer volume of data in such datasets poses significant hurdles in terms of computational resources and efficiency. As the dataset size increases, the number of granular-balls required to accurately represent the data grows exponentially. This not only demands more computational power but also increases the complexity of managing and analyzing these granular-balls.

In addition, the introduction of sampling-based acceleration techniques also invites new challenges. Selecting representative samples that accurately capture the inherent characteristics of the original dataset is crucial\cite{wu2023neighbourhood}. The performance and accuracy of the clustering results are significantly influenced by the chosen sampling strategy and sample size determination \cite{blanc2018adaptive}. Moreover, a trade-off exists between reducing computational complexity and potential information loss caused by sampling, thereby necessitating a balanced approach for efficiency and accuracy in these methods \cite{exarchakis2022sampling}.

To address these challenges, in this paper, a novel fast clustering algorithm is proposed, the main symbols used are listed in Appendix~A, and the major contributions of this paper are as follows: 

\begin{itemize}
\item \textbf{Novel Skeletal Data Representation}: We introduce a divide-and-conquer strategy that constructs multi-grained granular-balls from sample subsets to capture the essential topological skeleton of the data. This hierarchical representation preserves key clustering features while enabling efficient and scalable computation. 

\item \textbf{Linear-Time Clustering Algorithm}: Built upon this representation, we develop GBSK—a scalable clustering framework for large-scale data with $O(n)$ time complexity. The algorithm is governed by four parameters, ensuring efficient and effective clustering. 

\item \textbf{Simplified Adaptation for Practical Use}: To enhance usability, we introduce AGBSK, an adaptive variant that reduces parameter complexity to a single user-defined cluster count, $k$. This is made possible by employing  empirical  defaults for the number of sample sets, the sampling ratio, and the granular-ball count. 

\item \textbf{Systematic Scalability Verification}: Extensive experiments on various synthetic and real large-scale datasets—such as a 100-million-point dataset in 256 dimensions and datasets with nonspherical clusters—demonstrate the robustness of GBSK and show that it outperforms state-of-the-art methods on standard hardware. 
\end{itemize}

The rest of the paper is organized as follows. Section \ref{sec:related work} reviews the related work. 
In Section \ref{sec:preliminaries of gb}, we formalize the fundamental concepts of granular-ball computing. 
Section \ref{sec:GBSK} details our proposed method. Section \ref{sec:experiments} presents the experimental results and analysis. Finally, the conclusion is discussed in Section \ref{sec:conclusion}.

\section{Related Work}\label{sec:related work}
Recent studies on accelerating clustering algorithms have explored various techniques, each addressing specific challenges but often with limitations in broader applicability or scalability:

\textit{1) Acceleration based on granular-ball technology}:
Xia et al. \cite{xia2019granular} proposed the granular-ball model, where data are divided into granular-balls. Each granular-ball represents a localized region in the dataset, requiring only a center and a radius to describe the original data in any dimension. This approach significantly reduces computational requirements by using granular-balls instead of the entire dataset. 

The concept of granular-ball has been applied in supervised learning \cite{xia2024gbsvm, xia2022efficient, xia2021granular}. Granular-balls are also used in clustering. Xie et al. \cite{xie2023efficient} proposed a granular-ball based spectral clustering algorithm (GBSC), which modifies the method of constructing a similarity matrix. By utilizing granular-balls, the size of similarity matrix is significantly reduced. Cheng et al. \cite{cheng2023fast} proposed a granular-ball-based DPeak algorithm called GB-DP, which employs an unsupervised partitioning method to generate granular-balls from the original data. This approach replaces the size of the original data with a smaller number of balls, significantly reducing the running time of the DPeak algorithm. Xia et al. \cite{xia2022ballKmeans} proposed Ball k-means, which uses granular-balls for each cluster, yielding both higher performance and fewer distance calculations.

However, granular-ball technology has several limitations. 
First, its effectiveness relies heavily on the granularity level: excessively coarse granularity loses essential structural details, while overly fine granularity undermines computational efficiency.
Second, in high-dimensional spaces, the granular-ball method face trade-offs between accuracy and efficiency.  
Finally, as the dataset size increases, the number of granular-balls may grow quickly, leading to higher computational costs.

\textit{2) Acceleration based on sampling technology}:
k-means++ \cite{vassilvitskii2006k} accelerates k-means by initializing cluster centers through a sampling method. It selects random starting centers with specific probabilities, resulting in substantially improved performance. Recently, Ding et al. \cite{ding2023sampling} used an improved hybrid sampling \cite{huang2019ultra} method to select $p$ points from the original data, then accelerated the computation of density $\rho$ and relative distance $\delta$ for DPeak algorithm. Exarchakis et al. \cite{exarchakis2022sampling}  proposed a simple and efficient clustering method, named D-GM. It significantly improves efficiency by calculating the posterior over a specific subset which is iteratively refined by sampling in the neighbourhood of the best performing cluster at each EM iteration. 

Nonetheless, the clustering quality is highly dependent on the sample points, which may affect the stability of the clustering results. Inaccurate or biased sampling can lead to suboptimal cluster formation and may not accurately represent the underlying distribution of the dataset. 

\textit{3) Acceleration based on kNN technology}:
Acceleration based on kNN technique primarily relies on computing local density of each point by counting nearest neighbor number for it. The retrieval of k-nearest neighbors can be accelerated by employing spatial indexing structures, such as KD-tree, Cover-tree, FLANN \cite{muja2014scalable} and RP-forest.

FastDP \cite{sieranoja2019fast} introduces the kNN mechanism to compute the local density of points. FastDPeak \cite{chen2020fast} utilizes  cover-tree to speedup density computation, while Fast LDP-MST \cite{qiu2022fast} employs  KD-tree and RP-forest depending on dimensionality to improve the speed. These methods have an overall complexity of $O(nlogn)$ but are unsuitable for high-dimensional data. 

Yet, kNN-based acceleration algorithms typically rely heavily on spatial indexing structures, whose performance degrades significantly due to the curse of dimensionality. 

\textit{4) Acceleration based on parallel technology}:
Gong et al. \cite{gong2016eddpc} proposed EDDPC, which utilizes Voronoi partitioning with reasonable data replication and filtering to group data and process data objects in each group independently and in parallel. It then uses two complete MapReduce jobs to locally compute values within each grouping, avoiding a large number of unnecessary distance computations and data transfer overheads. EDDPC can achieve approximately 40 times performance improvement compared to simple MapReduce distribution. Zhang et al. \cite{zhang2016efficient} proposed the LSH-DDP algorithm, which partitions the input data using a locality-sensitive hashing (LSH) technique to increase the likelihood that the nearest neighboring points are assigned to the same partition. The algorithm performs local computations within each partition and aggregates these results to obtain an approximate final output. Compared to EDDPC, LSH-DDP performs similarly in terms of clustering results but doubles the processing speed. 

Still, acceleration based on parallel techniques may compromise the properties of the original data, potentially undermining the integrity of the clustering results. Moreover, the implementation relies on specific parallel models and software frameworks, which can complicate the deployment.

\textit{5) Acceleration based on grid technology}:
Wu et al. proposed the DGB\cite{wu2016fast} algorithm, which reduces computation time by partitioning the data space into fewer grid cells and replacing the original Euclidean distances between data points with distances between these non-empty cells. Similarly, the DPCG algorithm proposed by Xu \cite{xu2018dpcg} utilizes the grid concept from the CLIQUE algorithm to partition the data space, accelerating the computations for density and nearest neighbor distances for points with higher local density, with a time complexity of $O(m^2)$, where $m$ is the number of non-empty cells. PDPC \cite{xu2018fast}  employs grids to filter computations, quickly identifying dense regions. 

Despite this, in high dimension, both DGB and DPCG perform poorly or even degrade to $O(n^2)$.

\section{Granular-ball Method}\label{sec:preliminaries of gb}
The granular-ball method first treats the entire dataset in a coarse-grained manner, and then refines the data by splitting it using granular-balls, achieving a scalable and robust computational process.

\begin{definition}[granular-ball (ball for short)\cite{xia2019granular}]
    A granular-ball is a quintuple structure identified from a dataset $P$, defined as $ball=[E, c,  r, \rho, DM]$, where 
        $E \subset P$; 
        $c$ is the geometrical center of $E$; 
        $r$ is the radius of $ball$; 
        $\rho$ is the density of $ball$; 
        $DM$ is the distribution measure which is a weighted value. 
\end{definition}

Suppose in the dataset $P$, there are $t$ granular-balls that can be found, i.e., $ball_1, ball_2, \ldots, ball_t$. For each ball $ball_i$, its center $\mathrm{c}$ and radius $\mathrm{r}$ are defined as follows:
\begin{equation}\label{eq:ballCenter}
    ball_i.\mathrm{c} = \frac{1}{N_{ball_i}} \sum\limits_{j = 1}^{N_{ball_i}} {{x_{i,j}}}, 
\end{equation}
where $N_{ball_i}$ represents the number of points in $ball_{i}$, and $x_{i,j}\in ball_i$ denotes the $j$-th point within $ball_i$.
\begin{equation}\label{eq:ballRadius}
    ball_i.\mathrm{r} = \max_j ( \left\| x_{i,j} - ball_i.\mathrm{c} \right\|_2 )
\end{equation}

In spired by \cite{rodriguez2014clustering} \cite{cheng2023fast}, the density $\rho$ of $ball_i$ is defined as:
\begin{equation}\label{rho}
    ball_i.\rho = 
    \begin{cases} 
    \frac{N_{ball_i}}{ball_i.\mathrm{r} + medianR}, & \text{if } N_{ball_i} \neq 1\\
    0, & \text{if} N_{ball_i} = 1
    \end{cases}, 
\end{equation}
where $medianR$, a smoothing factor, is the median of $ball_1.\mathrm{r}$, $ball_2.\mathrm{r}$, $\ldots$, $ball_t.\mathrm{r}$. This factor plays a crucial role in balancing the density calculations, particularly for small granular-balls, ensuring that their size does not disproportionately influence the density estimation. 

The Distribution Measure $\mathrm{DM}$ is defined as follows:
\begin{equation}\label{eq:ballDM}
   ball_i.\mathrm{DM} = \frac{1}{{ball_i.\mathrm{r} + \tau}}, 
\end{equation}
where $\tau$ is a smoothing factor set to 0.01.

In the granular-ball method, $\mathrm{DM}$ is an important quantity used to split balls: Firstly, treat the whole dataset as a granular-ball $ball$. 
Secondly, split the $ball$ into two sub-balls $ball_L$ and $ball_R$, using k-means++ with $k=2$. 
Thirdly, the splitting of $ball$ succeeds if ${ball_L.\mathrm{DM}} < ball.\mathrm{DM}$ and $ball_R.\mathrm{DM} < ball.\mathrm{DM}$. 
 
\begin{figure}[thp]
    \centering
    \begin{subfigure}{0.23\textwidth}
        \centering
        \includegraphics[width=\linewidth]{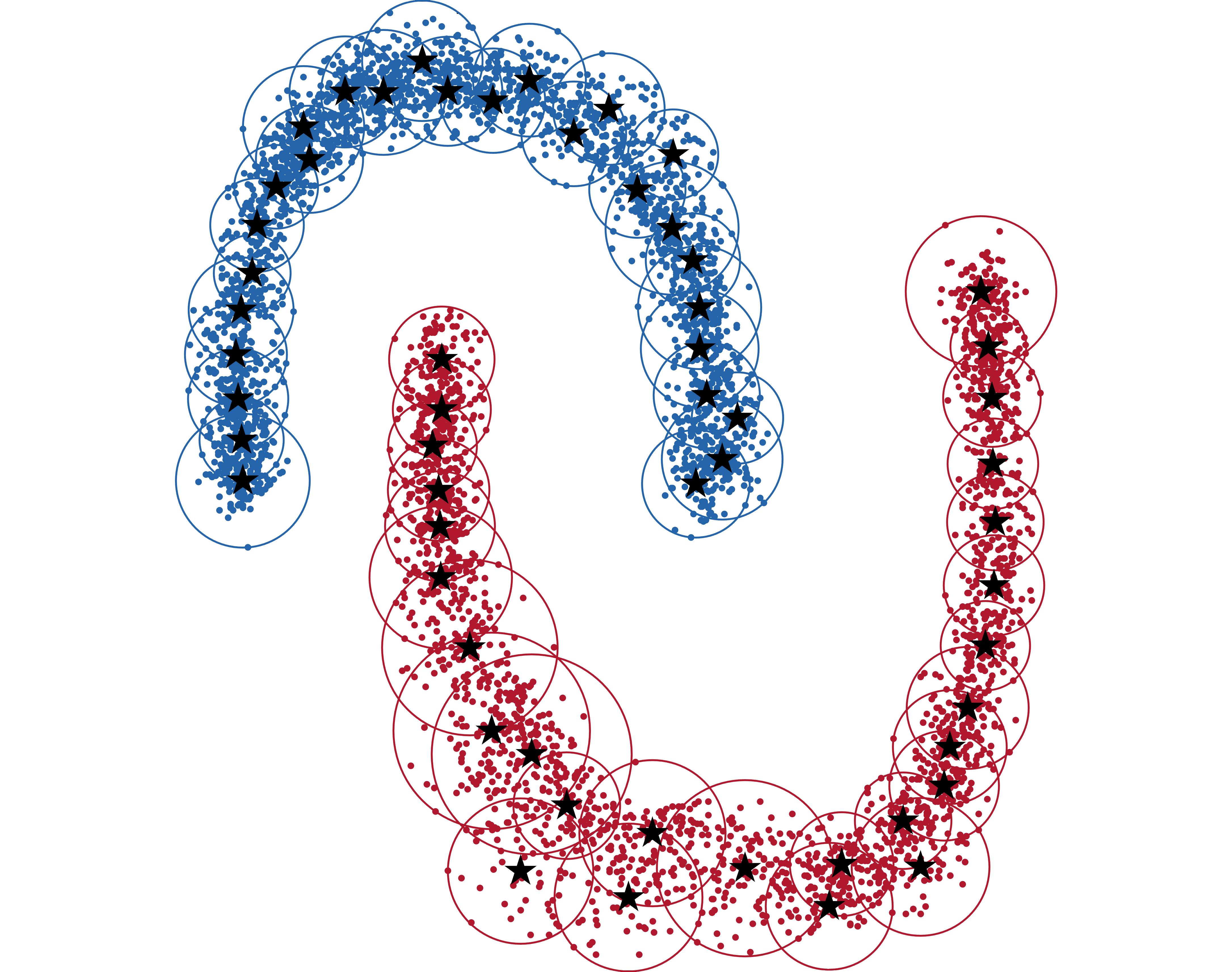}
        \caption{}
    \label{fig:banana_ball_1}
    \end{subfigure}
    \begin{subfigure}{0.23\textwidth}
        \centering
        \includegraphics[width=\linewidth]{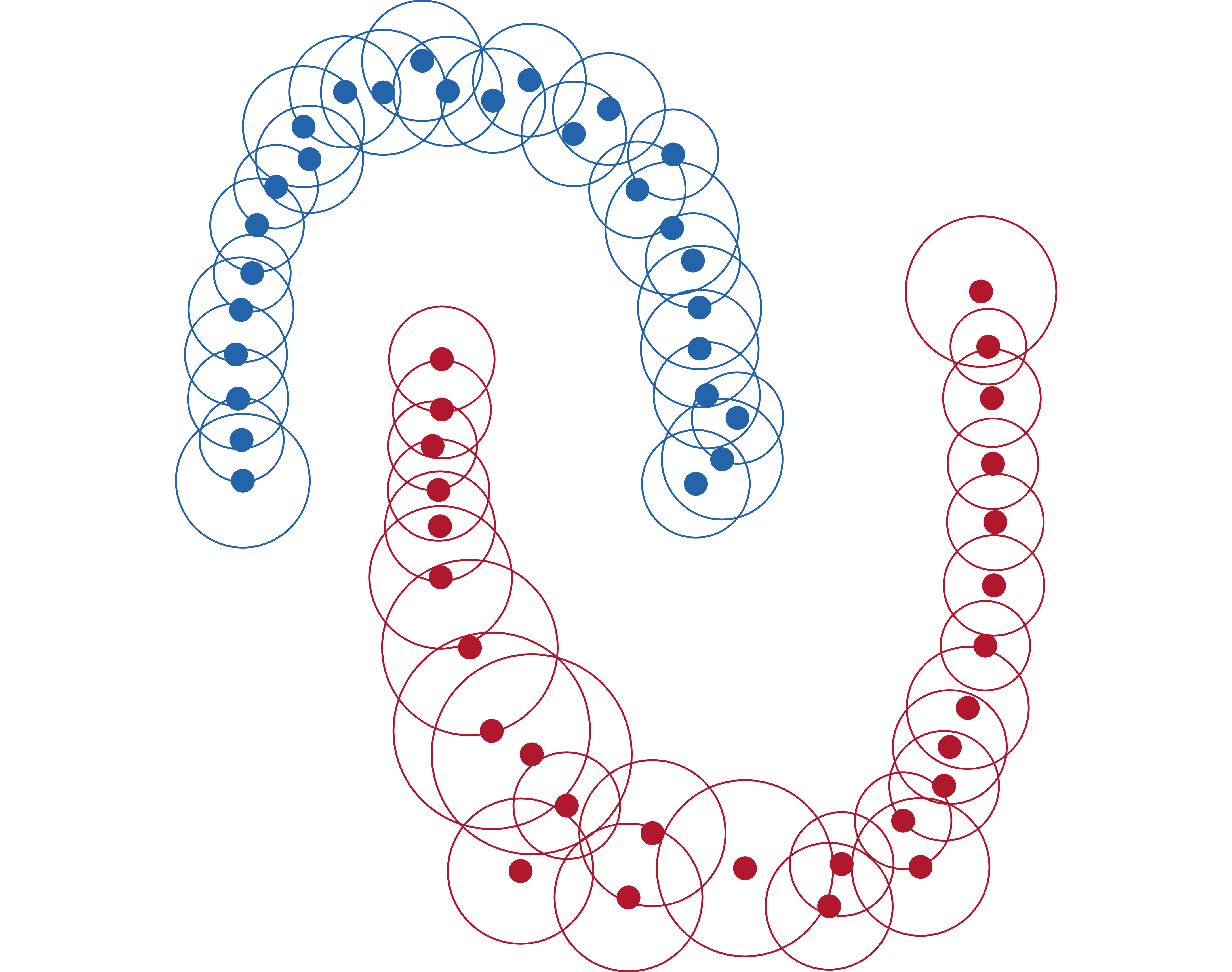}
        \caption{}
    \label{fig:banana_ball_2}
    \end{subfigure}
    \caption{\textbf{An example of granular-balls}: 
    (a) The granular-balls effectively cover the dataset's underlying structure; 
    (b) The data distribution of the granular-balls closely mirrors the original dataset's distribution.}
    \label{fig:banana_ball}
\end{figure}

\begin{algorithm}[thp]
\caption{Granular-ball generation}\label{alg:generateBalls}
\begin{algorithmic}[1] 
\REQUIRE $points$, number of granular-balls $M$
\ENSURE $balls$
\STATE push the granular-ball $gb$, which represents all points, into queue $Q_1$
\stepcomment{The loop generates $l \approx M$ balls}
\WHILE{$Q_1$ is not empty}
    \STATE $gb \gets$ pop a ball from $Q_1$
    \STATE compute $gb.\mathrm{c}$ and $gb.\mathrm{r}$ by \eqref{eq:ballCenter} and \eqref{eq:ballRadius}
    \STATE split $gb$ into two sub-balls $gb_L$ and $gb_R$ by k-means++
    \STATE compute $gb.\mathrm{DM}$, $gb_L.\mathrm{DM}$ and $gb_R.\mathrm{DM}$ by \eqref{eq:ballDM}
    \STATE compute $\operatorname{WDM}(gb)$ by \eqref{eq:ballWDM}
    \IF {$\operatorname{WDM}(gb) \geq gb.\mathrm{DM}$}
        \STATE push $gb_L$ and $gb_R$ into $Q_1$
    \ELSE
        \STATE $balls \gets balls \cup gb$
    \ENDIF
    \IF {$\operatorname{length}(balls) \geq M$ and $M \neq -1$}
        \STATE // $M=-1$: no restriction on the number of $balls$
        \STATE break
    \ENDIF
\ENDWHILE

\STATE $meanR \gets$ mean $r$ of all  balls in $balls$
\STATE $medianR \gets$ median $r$ of all balls in $balls$
\STATE push all balls into queue $Q_2$
\STATE $balls \gets \varnothing$
\stepcomment{The loop refines balls that are too big}
\WHILE{$Q_2$ is not empty}
    \STATE $gb \gets$ pop a ball from $Q_2$ 
    \IF {$gb.r \geq 2 \times max(meanR, medianR)$}
        \STATE split $gb$ into $gb_L$ and $gb_R$ by k-means++ 
        \STATE compute $gb_L.\mathrm{c}$ and $gb_R.\mathrm{c}$ by \eqref{eq:ballCenter}
        \STATE compute $gb_L.\mathrm{r}$ and $gb_R.\mathrm{r}$ by \eqref{eq:ballRadius}
        \STATE push $gb_L$ and $gb_R$ into $Q_2$
    \ELSE
        \STATE $balls \gets balls \cup gb$
    \ENDIF
\ENDWHILE
\RETURN $balls$
\end{algorithmic}
\end{algorithm}

\begin{figure*}[thp]
  \centering
  \includegraphics[width=0.95\textwidth]{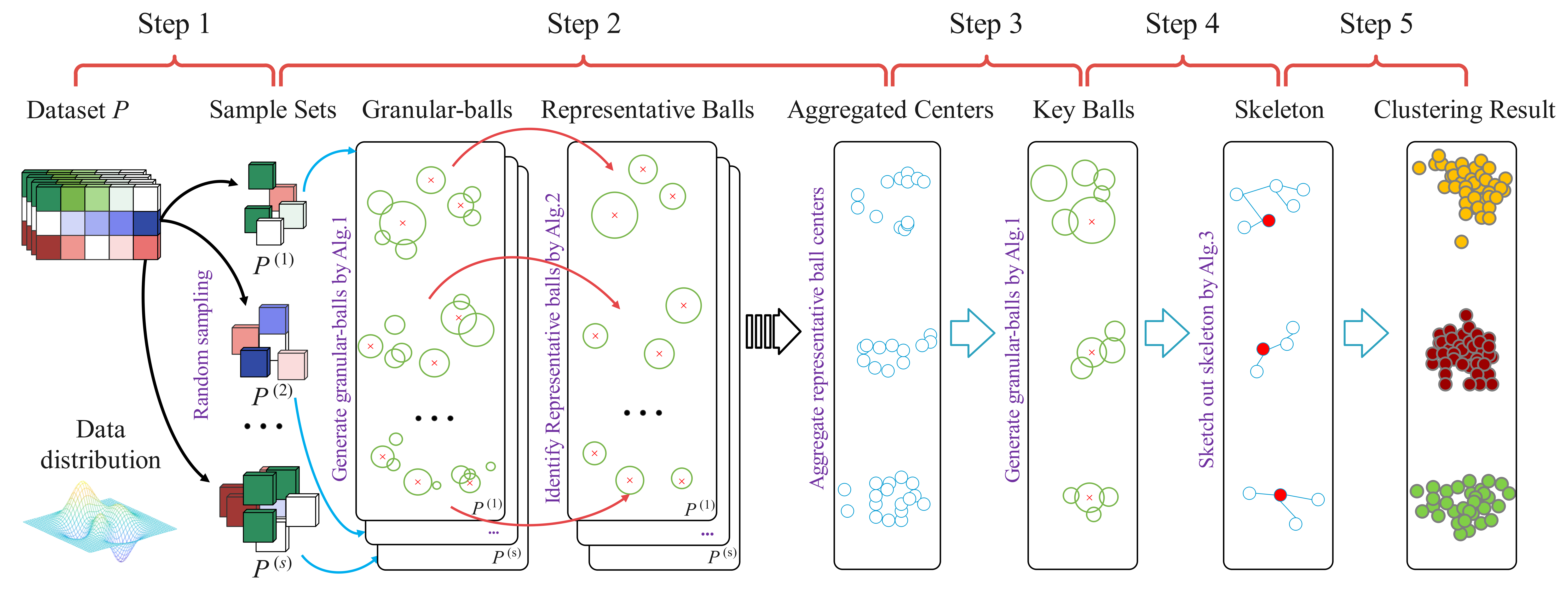}
  \caption{\textbf{Framework of GBSK}. In {Step 1}, we randomly obtain $s$ sample sets of the same size. 
  For each sample set, GBSK generates approximately $M$ number of granular-balls and identifies $k$ representative balls in {Step 2}. 
  In {Step 3}, GBSK constructs key balls from all representative balls. 
  {Step 4}, GBSK sketches out the skeleton by constructing a forest of key balls.  
  At last, it assigns labels to all remaining points based on the skeleton.}
  \label{fig:framework}
\end{figure*}

In noisy environments, the application of this rule may result in a proliferation of granular-balls. To address this challenge, the Weighted Decision Making ($\operatorname{WDM}$) value is introduced \cite{xie2023efficient}, which is designed to be more robust in noisy conditions:
\begin{definition}[Weighted Decision Making]
\begin{equation}\label{eq:ballWDM}
    \operatorname{WDM}(ball) = \frac{N_{ball_L}}{N_{ball}}(ball_L.\mathrm{DM}) + \frac{N_{ball_R}}{N_{ball}}(ball_R.\mathrm{DM}).
\end{equation}
\end{definition}

Fig.~\ref{fig:banana_ball} demonstrates how the granular-balls can cover the dataset while maintaining a similar distribution. Algorithm \ref{alg:generateBalls} delineates the procedure of generating granular-balls from a dataset, comprising two main stages. The initial stage involves generating $balls$, followed by a refinement stage for $balls$ that exceed a specified size. The parameter $M$ plays a crucial role in controlling the approximate number of $balls$: if $M>0$, the actual number of $balls$ returned closely approximates $M$; conversely, setting $M=-1$ imposes no numerical restrictions on the number of $balls$.

\section{GBSK: The Proposed Method}\label{sec:GBSK}
\subsection{Motivation}
The motivation for developing GBSK arises from the difficulties associated with clustering large-scale datasets, particularly those whose distribution is unknown. Our hypothesis posits that independent of data distribution, there is an inherent ``skeleton'', which is a fundamental structure encapsulating the essential characteristics and patterns of the data. 

Therefore, by focusing on uncovering this skeleton, we can leverage structural insights to guide the clustering process, substantially reducing the computational load and enhancing the speed of clustering. 

\subsection{Methodology}
First, granular-balls are a multi-grained representation of data that mirrors a distribution similar to the original data \cite{xia2019granular}. Henceforth, the granular-ball method is suitable for depicting the data backbone, i.e., the core structure or essence of the dataset, thereby speedup the classification process. 

Second, according to extensive experiments, it is observed that multiple samples from the original dataset reveal statistical regularities, which reflect the distribution of the entire dataset. Consequently, we posit the following hypothesis.

In a single small-scale random sample set, clustering can yield outcomes with intrinsic randomness, potentially leading to a considerable deviation from the distribution of the original dataset $P$. However, a certain statistical regularity is expected to emerge, if the sampling process is repeated sufficiently enough, generating a series of sample sets denoted as $SP = \{P^{(1)}, P^{(2)}, \ldots\}$, where $P^{(i)}$ stands for the $i$-th sample set of $P$. This regularity suggests that the distribution of $SP$ will approximate the overall distribution of $P$. 

Hence, it is advantageous for leveraging the divide-and-conquer strategy to uncover the data skeleton by randomly sampling the original dataset $P$ multiple times into smaller sets. These sample sets can be represented by multi-grained granular-balls, encapsulating the data distributions and spatial structures within each set. 

Specifically, the data skeleton emerges progressively through iteratively granular-ball generation across adequate small-scale sample sets. This process constructs a multi-scale spatial representation that captures the dataset's essence through successive refinements, and that converges toward a stable topological representation of the original data distribution. The skeleton refer to an evolving approximation rather than a single static entity. 

\subsection{Framework of GBSK}\label{subsec:framework}

\begin{figure*}[th]
    \centering

    \begin{subfigure}[b]{0.155\textwidth}
        \includegraphics[width=\textwidth]{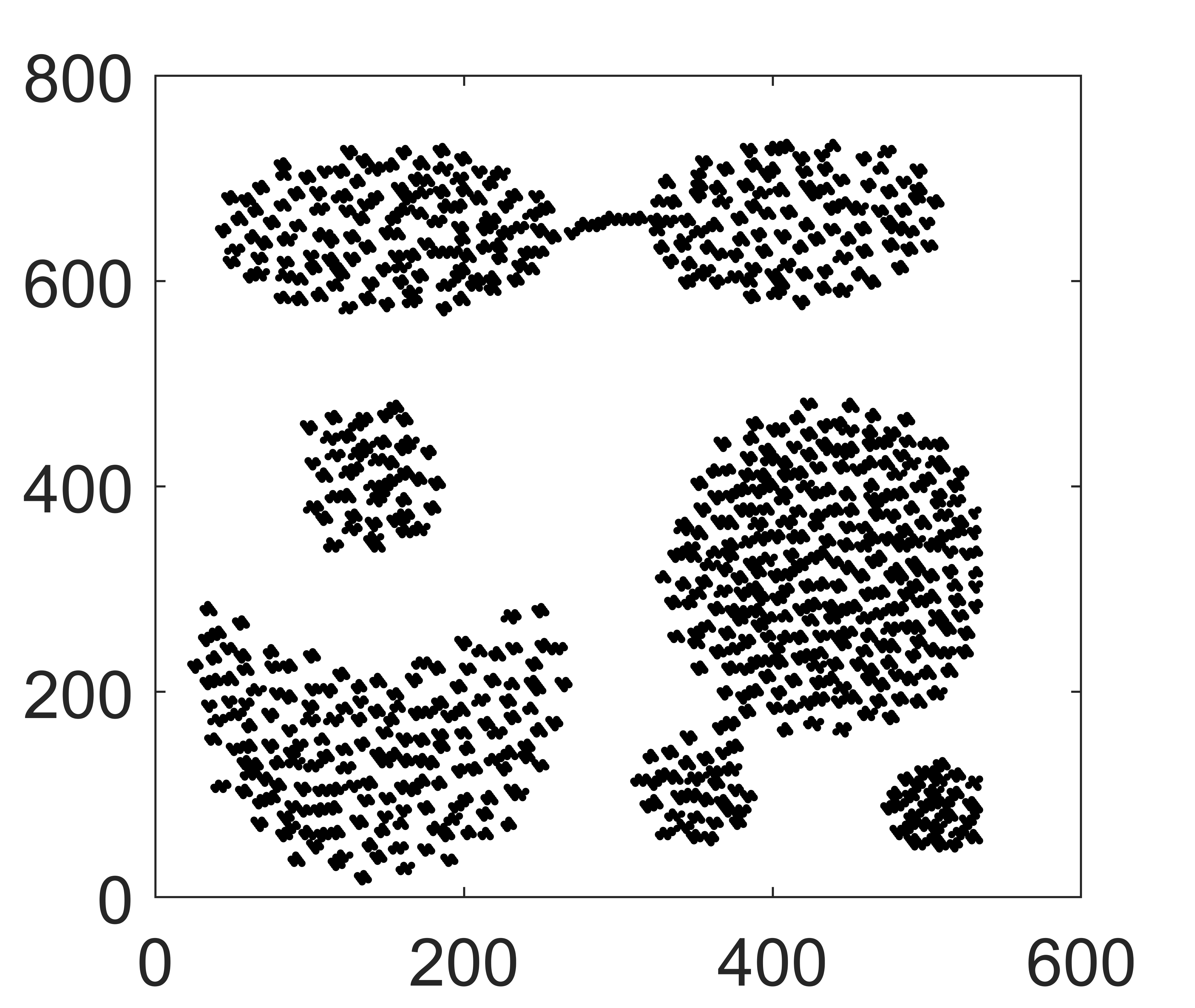}
        \caption{Data (\textsc{SYN2})}
    \end{subfigure}
    \begin{subfigure}[b] {0.155\textwidth}
        \includegraphics[width=\textwidth]{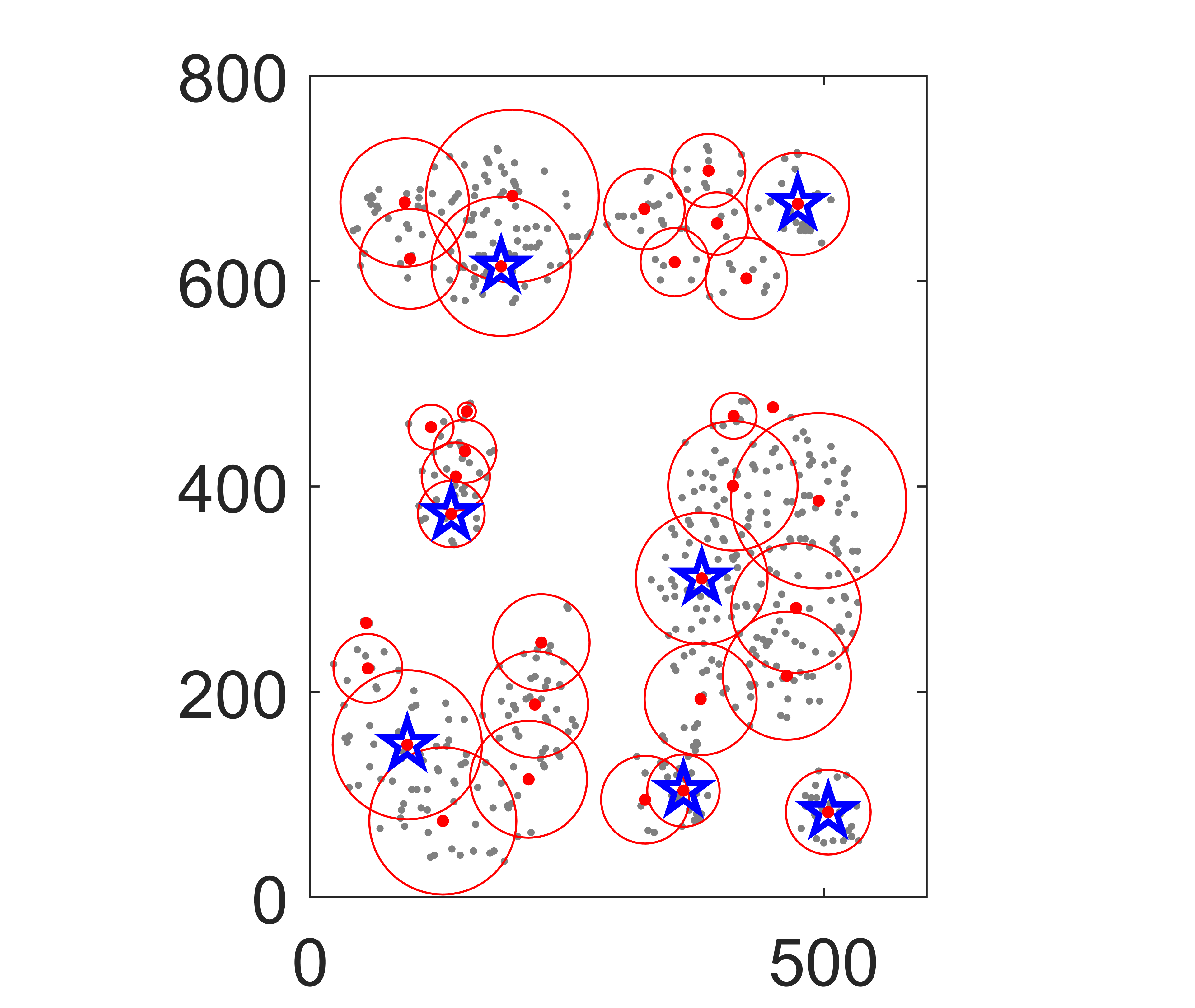} 
        \caption{Sample $P^{(1)}$}
        \label{fig:sampleP1_SYN2}
    \end{subfigure}
    \begin{subfigure}{0.01\textwidth}
        \centering
        \raisebox{0.08\textheight}[0pt][0pt]{\hspace*{-5.5pt}\textbf{\ldots}}
    \end{subfigure}
    \vspace*{\fill}
    \begin{subfigure}[b]{0.155\textwidth}
        \includegraphics[width=\textwidth]{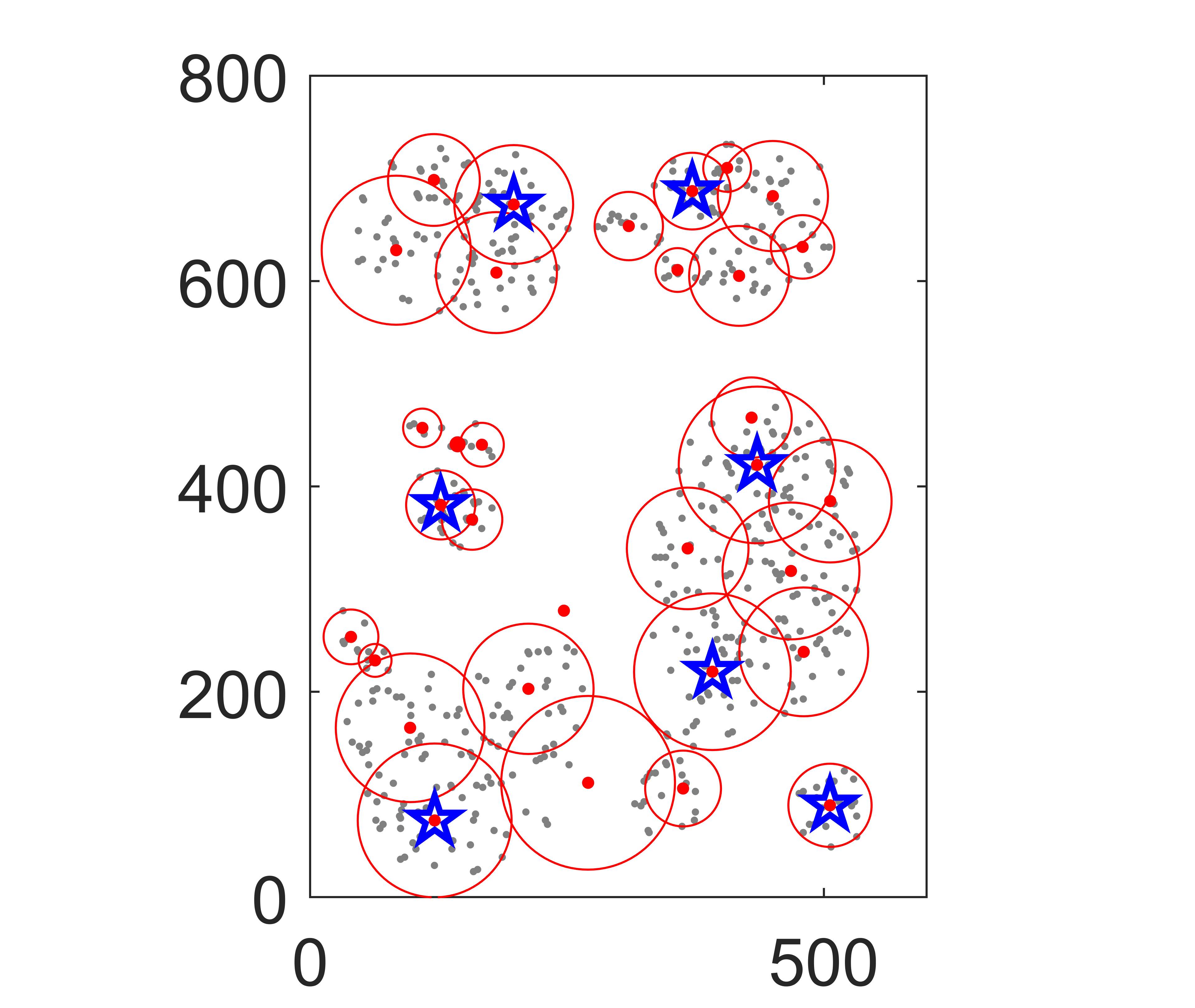}
        \caption{Sample $P^{(s)}$}
        \label{fig:samplePs_SYN2}
    \end{subfigure}
    \begin{subfigure}[b]{0.155\textwidth}
        \includegraphics[width=\textwidth]{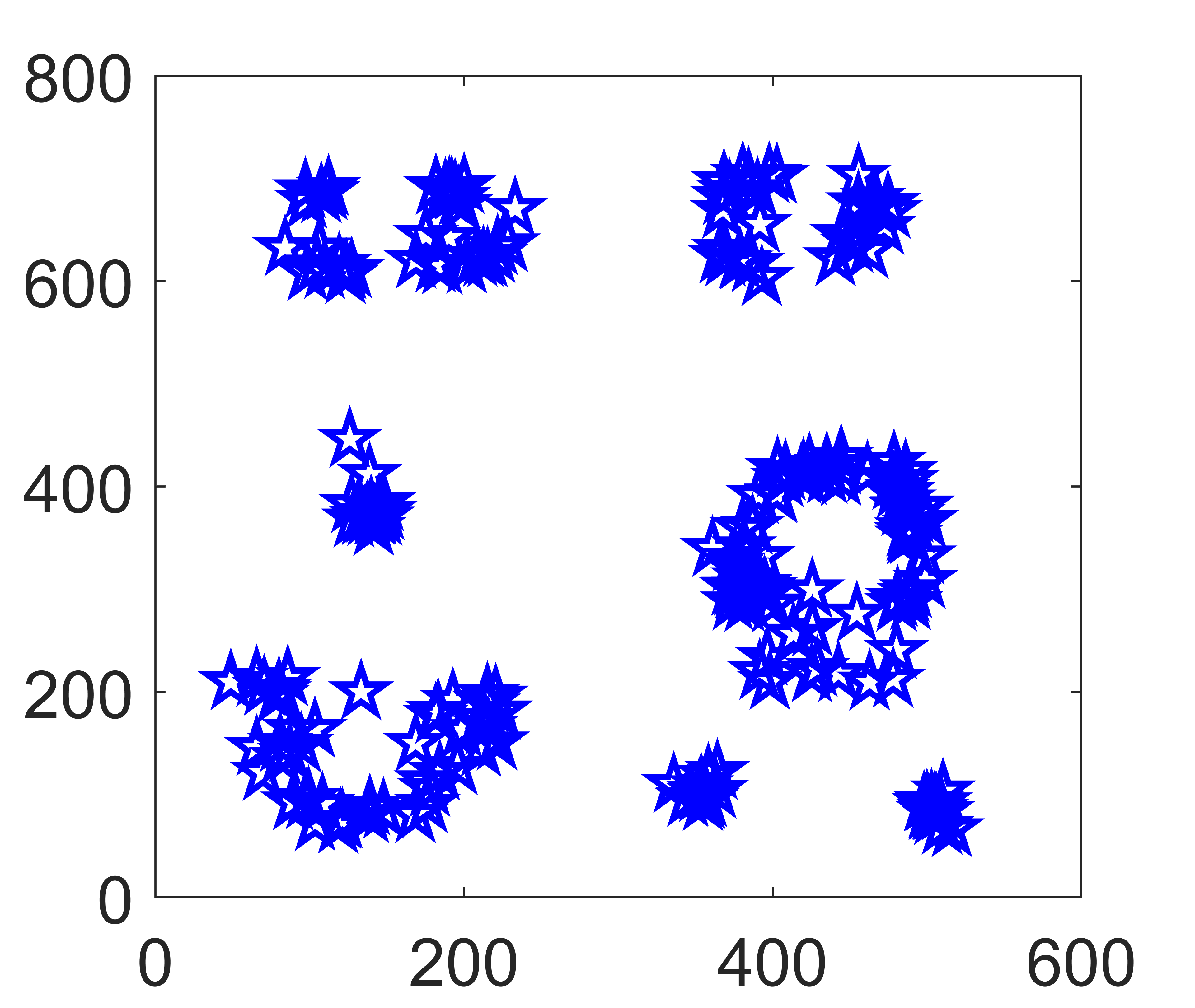}
        \caption{ARBC}
        \label{fig:ARBC_SYN2}
    \end{subfigure}
    \begin{subfigure}[b]{0.155\textwidth}
        \includegraphics[width=\textwidth]{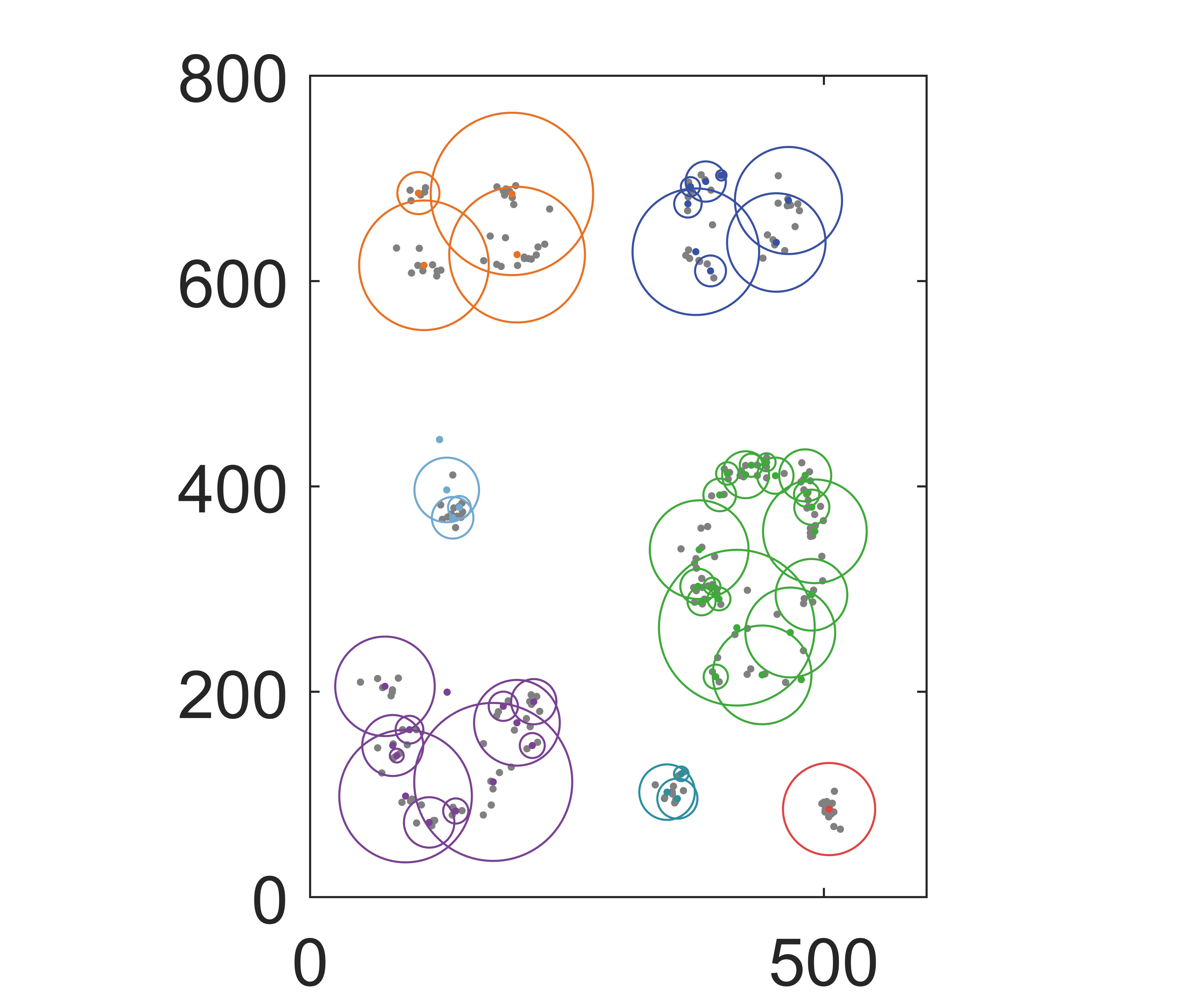}
        \caption{Key balls}
        \label{fig:KEYBALLS_SYN2}
    \end{subfigure}
    \begin{subfigure}[b]{0.155\textwidth}
        \includegraphics[width=\textwidth]{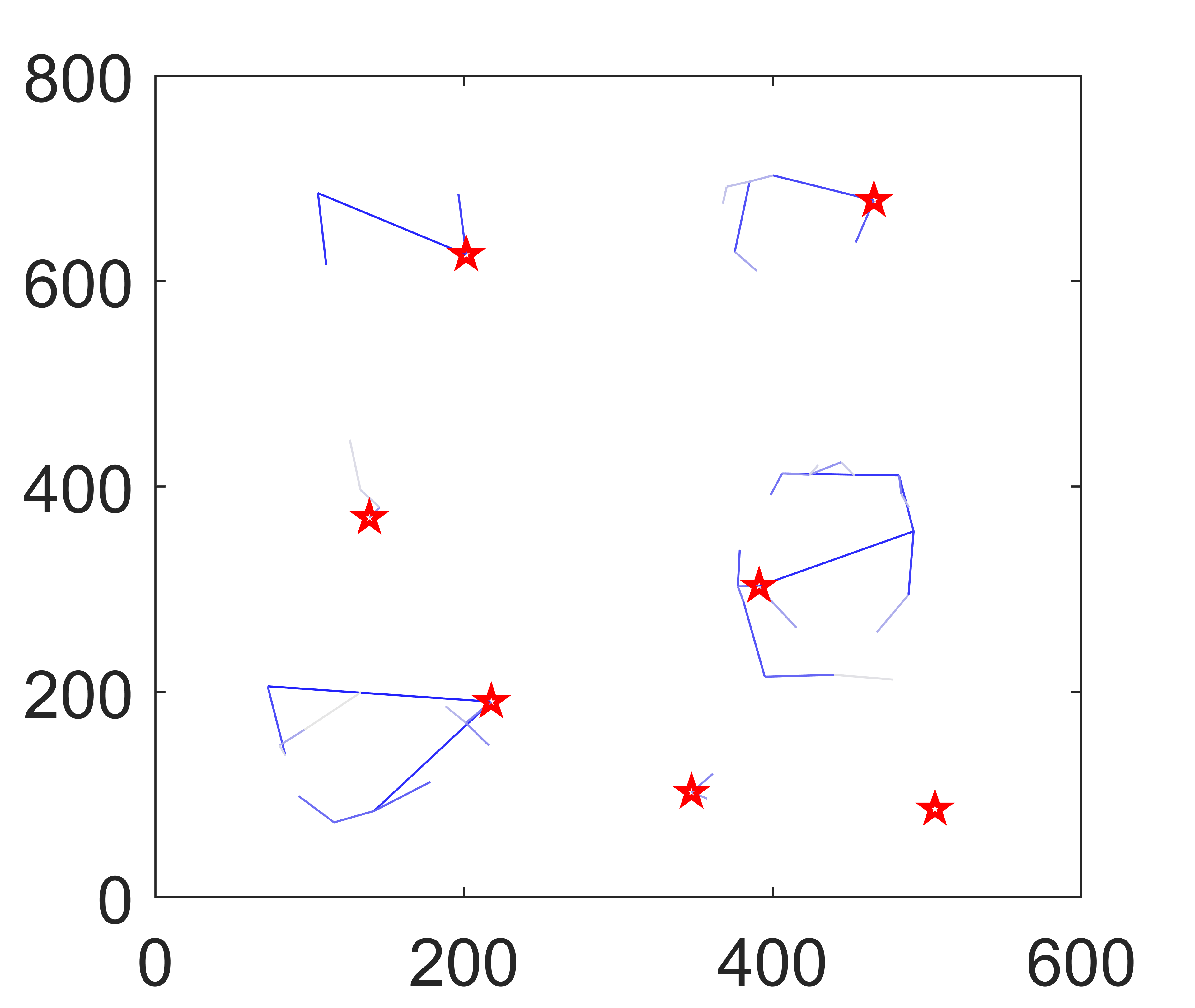}
        \caption{Skeleton}
        \label{fig:Skeleton_SYN2}
    \end{subfigure}

    \caption{An example of the \textbf{skeleton construction process} of GBSK on \textsc{SYN2}. 
    (a) Distribution of \textsc{SYN2}; 
    (b)-(c) $s$ sample sets, where red circles represents granular-balls and blue stars denote the centers of representative balls; 
    (d) Aggregated centers of representative balls across all sample sets; 
    (e) Key balls identified from (d); 
    (f) Resulting skeleton.}
    \label{fig:SYN2SkeletonConstruction}
\end{figure*}

\begin{figure*}[th]
    \centering
    \includegraphics[scale=0.4]{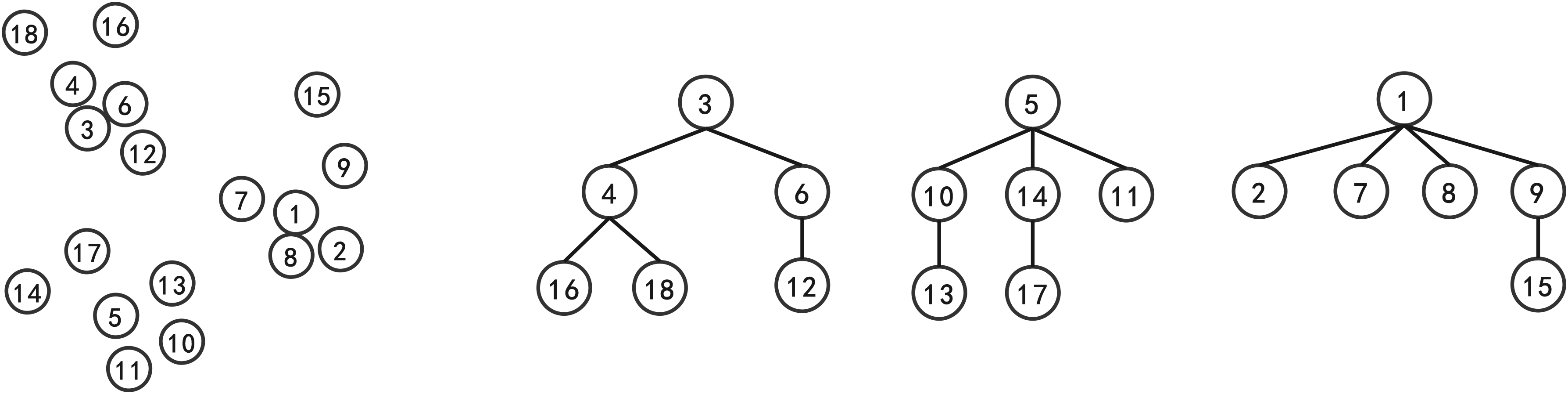}
    \caption{An example of \textbf{building a forest} with three trees for a dataset. On the left, the distribution of the dataset is shown. On the right, three trees are constructed according to the principle that each point has a unique parent, which is its nearest neighbor with higher density. The numbers in the data points indicate their descending order of density.}
    \label{fig:forestConstruction}
\end{figure*}

Inspired by the thoughts above, the skeleton clustering is proposed, which divides a large dataset into small subsets and is characterized by multi-grained granular-balls. Given a $d$-dimension dataset $P$, the main processing steps of GBSK are exhibited as seen in Fig.~\ref{fig:framework}. 

In brief, the GBSK algorithm follows a tightly integrated process, of which each step builds on the previous to ensure efficient and scalable clustering of large-scale datasets. It begins with Step 1, in which random sampling reduces computational load, providing manageable subsets for granular-balls construction. In Step 2, for each sample set, granular-balls are identified, from which representative balls are picked to summarize local structures. Step 3, generate key granular-balls on representative ball centers, minimizing redundancy while encapsulating the intrinsic features and global structure of the dataset $P$. Subsequently, Step 4 organizes the key balls into a hierarchical forest structure, representing the skeleton of $P$ and offering a clear framework for clustering. Finally, in Step 5, this skeleton guides the final clustering process, allowing the entire dataset to be efficiently labeled. 

Each step systematically abstracts and progressively refines the dataset, transitioning from local to global structure, thereby enabling precise and efficient large-scale clustering. See below for details.

\textbf{Step 1 (Random Sampling)}: Obtain sample sets $P^{(1)}$, $P^{(2)}$, $\ldots$, $P^{(s)}$ by sampling from $P$ randomly and repeatedly enough. $s$ is the number of sample sets. Each sample set $P^{(i)}=\{p^{(i)}_1, p^{(i)}_2, \ldots, p^{(i)}_j, \ldots, p^{(i)}_{sampleSize}\}$, $1 \le i \le s$, contains $sampleSize=n\times \alpha$ points. $p^{(i)}_j$ is the $j$-th point. An example of sample sets generated from the \textsc{SYN2} dataset is shown in Fig.~\ref{fig:sampleP1_SYN2} and Fig.~\ref{fig:samplePs_SYN2}.

\textbf{Step 2 (Identifying Representative Balls)}: Similar to the strategy employed by DPeak to find peaks, we identify $k$ dense representatives for $k$ clusters. 

For each sample set, use Algorithm \ref{alg:generateBalls} to generate $l$ granular-balls, namely $ball_1^{(i)}$, $ball_2^{(i)}$, \ldots, $ball_l^{(i)}$, where $l \approx M$. For each $ball_j^{(i)}$, $1 \le j\le l$, compute $\delta_j^{(i)}$, the distance of $ball_j^{(i)}$ to its nearest neighbor granular-ball with higher density in $P^{(i)}$: 
\begin{equation}\label{eq:delta}
\delta_{j}^{(i)}=\min_{q:ball_{q}^{(i)}.\rho>ball_{j}^{(i)}.\rho} \operatorname{ballDist}(ball_q^{(i)}, ball_j^{(i)}),
\end{equation}
where $\operatorname{ballDist}(ball_q^{(i)},ball_j^{(i)})=\left\|ball_q^{(i)}.\mathrm{c}-ball_j^{(i)}.\mathrm{c}\right\|_2$. Then, compute $\gamma_j^{(i)}$ for selecting peaks by: 
\begin{equation}\label{eq:gamma}
    \gamma_j^{(i)}=\rho_{j}^{(i)} \times \delta_j^{(i)}. 
\end{equation}

\begin{algorithm}[t]
    \caption{Identifying peak (or representative) balls}  
    \label{alg:identifyPeakBalls}
    \begin{algorithmic}[1] 
        \REQUIRE $balls$, category number $k$ 
        \ENSURE $peakBalls$ 
        \FOR{each $ball_j \in balls$}
            \STATE compute $\rho_{j}$ by \eqref{rho}
            \STATE compute $\delta_{j}$ by \eqref{eq:delta}
            \STATE compute $\gamma_{j}$ by \eqref{eq:gamma}
        \ENDFOR
        \RETURN $peakBalls =$ top $k$ balls with largest $\gamma$ value
    \end{algorithmic}
\end{algorithm}

    

\begin{algorithm}[th] 
    \caption{Constructing a forest for key balls}
    \label{alg:constructForest}
    \begin{algorithmic}[1] 
        \REQUIRE $KEYBALLS$, number of trees (category number) $k$
        \ENSURE $parentIDs$, $rootIDs$ // forest structure
        \STATE $parentIDs = \{-1, -1, \ldots, -1\}$ //parent indexes
        \STATE $rootBalls =$ identify $k$ balls by Algorithm \ref{alg:identifyPeakBalls} from $KEYBALLS$
        \FOR{each $keyBall_i \in KEYBALLS$}
            \IF{$keyBall_i \notin rootBalls$}
                \STATE $parentIDs(i) \gets$ parent index of $keyBall_i$ by \eqref{eq:findParentBall}
            \ENDIF
        \ENDFOR
        \STATE // ``$-1$'' indicates the root of a tree
        \STATE  $rootIDs= \{j|parentIDs(j) == -1\} $
        \RETURN $parentIDs$, $rootIDs$
    \end{algorithmic}
\end{algorithm}

\begin{itemize}
    \item After repeatedly sampling within a large dataset, the number of granular-balls may still be substantial, leading to considerable computational overhead. 
    \item These representative balls effectively capture the most essential local information by preserving structural relationships and distribution characteristics while filtering out noise.  
\end{itemize}

\begin{definition}[Representative Balls]
    The granular-balls $\{ball_1^{(i)}$, $ball_2^{(i)}$, $\ldots$, $ball_l^{(i)}\}$, identified from $P^{(i)}$, are sorted in descending order of $\gamma$. The top k balls with largest $\gamma$ values are called \textbf{Representative Balls}, $repBalls^{(i)} = $ $\{repBall_1^{(i)}$, $repBall_2^{(i)}$, $\ldots$, $repBall_k^{(i)}\}$, covering the high-density regions of each localized region in the sample set. 
\end{definition}

Algorithm \ref{alg:identifyPeakBalls} outlines the procedure for identifying representative balls. Using representative balls rather than all granular-balls is motivated by two considerations. 

Hence, by focusing on representative granular-balls, we can not only reduce the computational complexity but also retain essential information required for accurate clustering. 

Let $repBallCenters^{(i)}$ $= \{$$repBall_j^{(i)}.\mathrm{c}$, $1 \le j \le k\}$ be the center set of all representative balls in $P^{(i)}$. Let $aggRepBallCenters$ $= repBallCenters^{(1)}$ $\bigcup$ $repBallCenters^{(2)}$ $\bigcup$ $\ldots$ $\bigcup$ $repBallCenters^{(s)}$ be the aggregated center set from all sample sets. We refer to $repBallCenters^{(i)}$ as Local Representative Centers and $aggRepBallCenters$ as Aggregated-Representative Centers 
(\textbf{ARBC}) (e.g. Fig.~\ref{fig:ARBC_SYN2}). 

\textbf{Step 3 (Identifying Key Balls)}: The granular-balls identified by Algorithm \ref{alg:generateBalls} on $aggRepBallCenters$ are called \textbf{Key Balls}, $KEYBALLS$ $= \{keyBall_j, 1 \le j \le W\}$, where $W \leq k \times s$ (e.g. Fig.~\ref{fig:KEYBALLS_SYN2}). 

\textbf{Step 4 (Sketching Out the Skeleton)}: DPeak constructs a forest where each tree reflects the structure of a category in the dataset \cite{chen2020fast} (e.g., Fig.~\ref{fig:forestConstruction}). Similarly, we build a forest over the set of key granular-balls $KEYBALLS$. For each  $keyBall_i \in KEYBALLS$, its parent ball is selected as the nearest key ball with higher density. 
\begin{equation}\label{eq:findParentBall}
    \begin{split}
        &\operatorname{ParentID}(keyBall_i) = \\
        &\underset{j:keyBall_j.\rho>keyBall_i.\rho}{\mathrm{argmin}} \operatorname{ballDist}(keyBall_i, keyBall_j).
    \end{split}
\end{equation}

Algorithm \ref{alg:constructForest} presents the main procedure of constructing a forest for $KEYBALLS$. We refer to this forest as the \textbf{skeleton} of the entire dataset $P$, where each tree within the forest corresponds to a distinct category (e.g. Fig.~\ref{fig:Skeleton_SYN2}). 

\begin{algorithm}[th]
    \caption{Granular-ball Skeleton Clustering (GBSK)}
    \label{alg:GBSK}
    \begin{algorithmic}[1] 
        \REQUIRE dataset $P$ ($n \times d$ matrix), number of sample sets $s$, sampling proportion $\alpha$, granular-balls number $M$, category number $k$ 
        \ENSURE clustering labels of points in $P$
        
        \stepcomment{Step 1. Random Sampling}
        \STATE $sampleSize = n \times \alpha$ // sample set size
        \STATE obtain sample sets $P^{(1)}$, $P^{(2)}$, \ldots, $P^{(s)}$ from $P$
        
        \stepcomment{Step 2. Identifying Representative Balls}
        \STATE $aggRepBallCenters \gets \varnothing$
        \FOR {each $P^{(i)}$}
             \STATE $balls^{(i)} \gets$ find $l\approx M$ balls by Algorithm \ref{alg:generateBalls}
             \STATE $repBalls^{(i)} \gets$ $k$ representative balls by Algorithm \ref{alg:identifyPeakBalls}
             \STATE $repBallCenters^{(i)} =\{ball.c| ball\in repBalls^{(i)}\}$ 
             \STATE push $repBallCenters^{(i)}$ into $aggRepBallCenters$
        \ENDFOR
        
        \stepcomment{Step 3. Identifying Key Balls}
        \STATE $M = -1$ // number of key balls is not limited
        \STATE $KEYBALLS =$ obtain $W$ key balls by Algorithm \ref{alg:generateBalls} on $aggRepBallCenters$

        \stepcomment{Step 4. Sketching Out the Skeleton}
        \STATE $[rootIDs, parentIDs]=$ build a forest by Algorithm \ref{alg:constructForest} 
        \FOR{$ i=1~to ~|rootIDs|$}
            \STATE $j= rootIDs(i) $ // index of root key ball
            \STATE assign a new label to the $j$-th ball in $KEYBALLS$
        \ENDFOR
        \FOR{$\forall keyBall_i \in KEYBALLS$}
            \STATE assign a label to $keyBall_i$ by \eqref{eq:labelNonRootKeyBall}
        \ENDFOR
       
        \stepcomment{Step 5. Final Clustering}
        \FOR{each $p \in P$}
            \STATE assign a label to $p$ by \eqref{eq:labelPoints}
        \ENDFOR
    \end{algorithmic}
\end{algorithm}

All key balls within the same tree are assigned a common label. Specifically, each root key ball is assigned a unique label, and each key ball $keyBall_i \in KEYBALLS$ inherits the label of its corresponding root node: 
\begin{equation}\label{eq:labelNonRootKeyBall}
     \operatorname{label}(keyBall_i) = \operatorname{label}(\operatorname{root}(keyBall_i)), 
\end{equation}
where $\operatorname{root}(keyBall_i)$ returns the root node of $keyBall_i$ within the forest. 

\textbf{Step 5 (Final Clustering)}: Each unclassified point $p \in P$ is directly assigned to the category of its nearest key ball: 
\begin{equation}\label{eq:labelPoints}
    \begin{cases}
        \operatorname{nearestBall}(p) = \underset{ball \in KEYBALLS}{\mathrm{argmin}} \|p-ball.\mathrm{c}\| \\
        \operatorname{label}(p) = \operatorname{label}(\operatorname{nearestBall}(p))
    \end{cases}. 
\end{equation}

Algorithm \ref{alg:GBSK} presents the proposed method. 

\subsection{Parameter Setting for GBSK}\label{subsec:parameters}
The original GBSK algorithm includes four tunable parameters, offering flexibility but potentially challenging for beginners to configure. Parameter tuning—especially on large-scale or high-dimensional datasets—can be both time-consuming and computationally intensive. To enhance usability without compromising performance, we provide empirically validated default settings for $s$, $\alpha$, and $M$, derived from extensive experimentation and analysis: 

\begin{figure*}[tbh]
    \centering
    
    \begin{subfigure}{0.245\textwidth}
        \centering
        \includegraphics[scale=0.245]{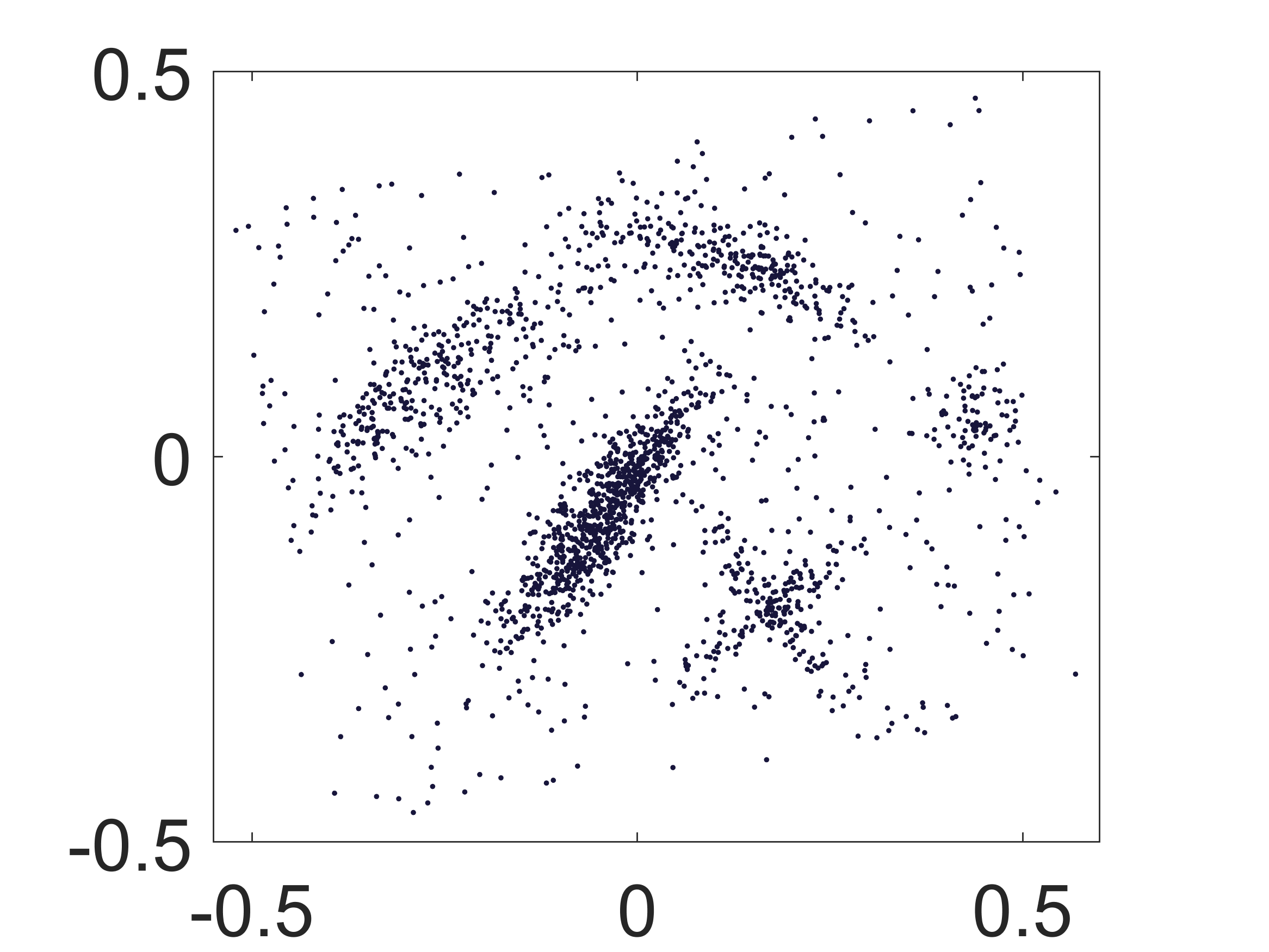}
        \caption{\textsc{SYN1}}
        \label{SYN1:data}
    \end{subfigure}
    \begin{subfigure}{0.49\textwidth}
        \centering
        \includegraphics[scale=0.245]{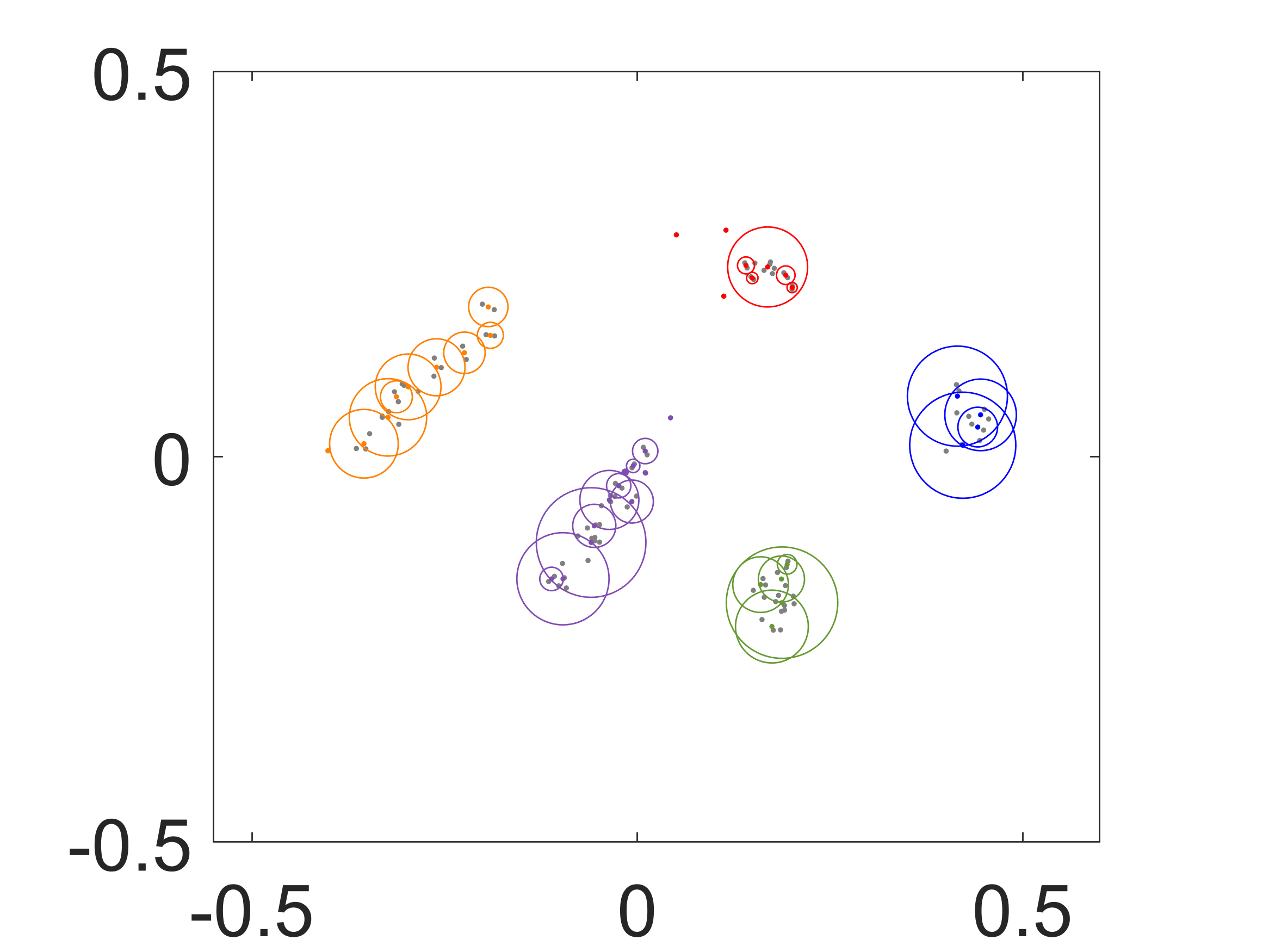}
        \includegraphics[scale=0.245]{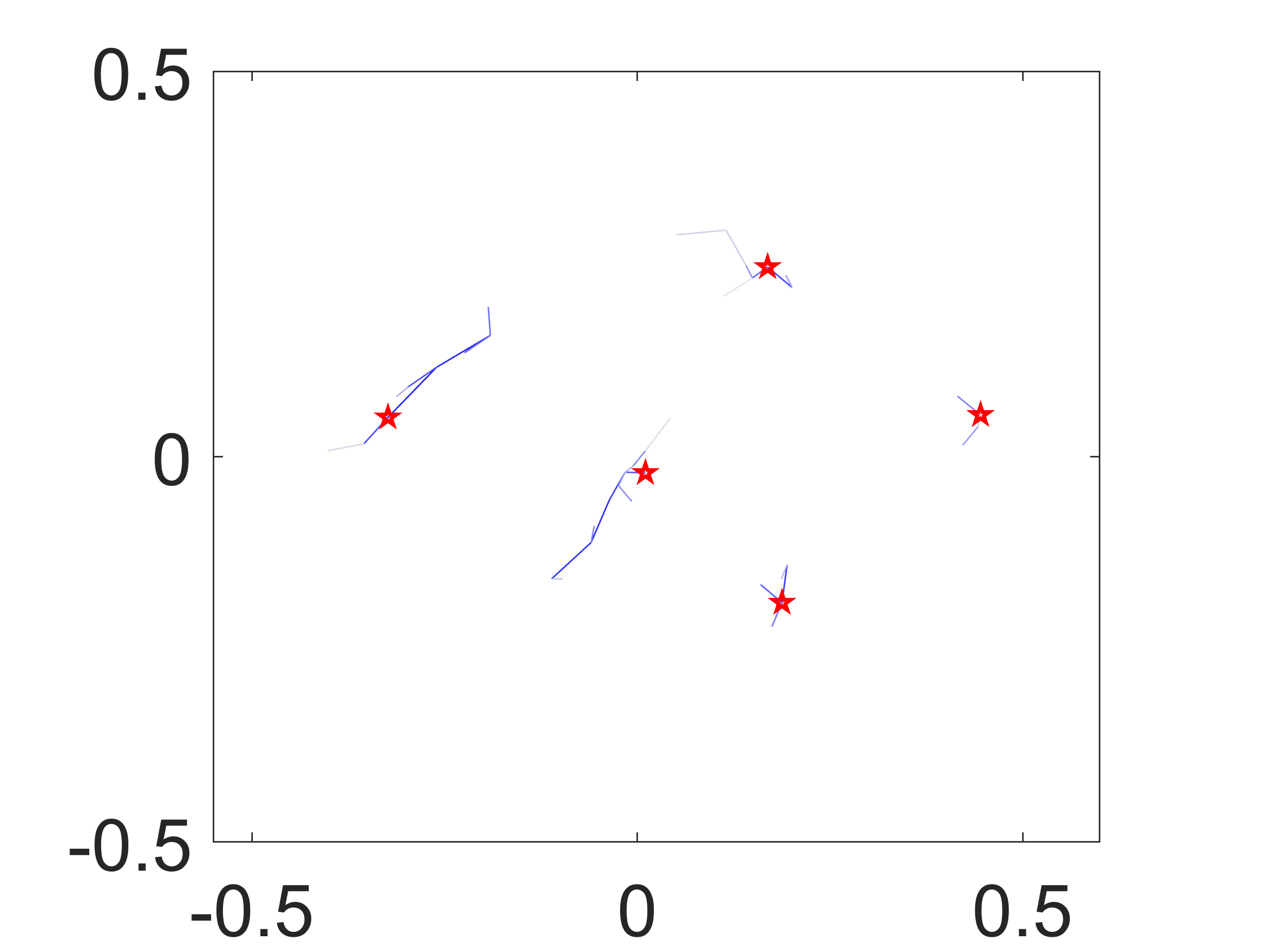}
        \caption{Key granular-balls and skeleton}
        \label{SYN1:skeleton_and_label}
    \end{subfigure}
    \begin{subfigure}{0.245\textwidth}
        \centering
        \includegraphics[scale=0.245]{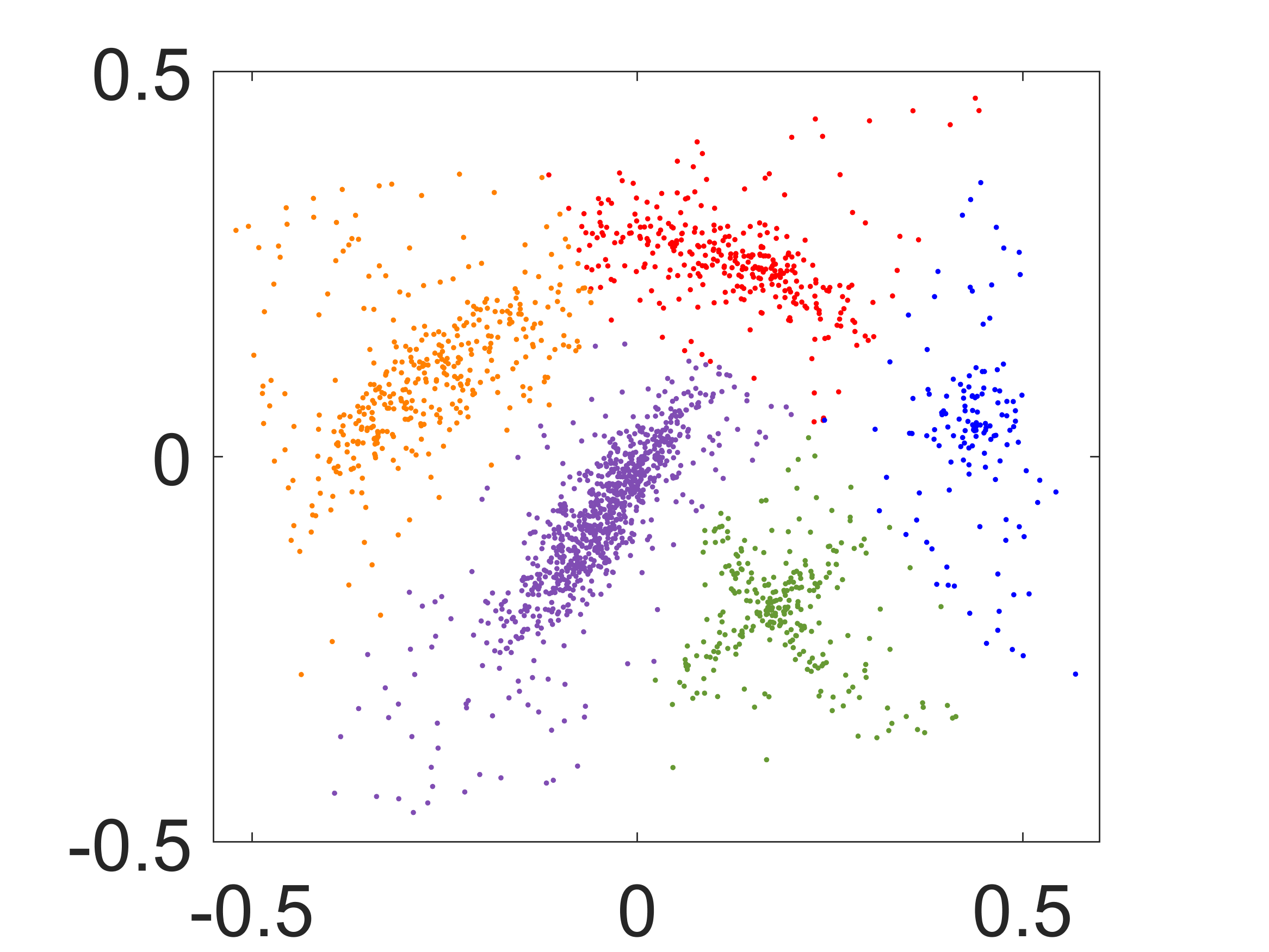}
        \caption{Result}
        \label{SYN1:DP_label}
    \end{subfigure}
    
    \caption{A 2D example: 
    (a) Original data of \textsc{SYN1}. 
    (b) The left shows the key balls, and the right shows the skeleton of \textsc{SYN1}. 
    (c) Clustering result of GBSK.}
    \label{fig:SYN1}
    \vspace{-9pt} 
\end{figure*}

\begin{figure*}[tbh]
    \centering
    
    \begin{subfigure}{0.245\textwidth}
        \centering
        \includegraphics[scale=0.245]{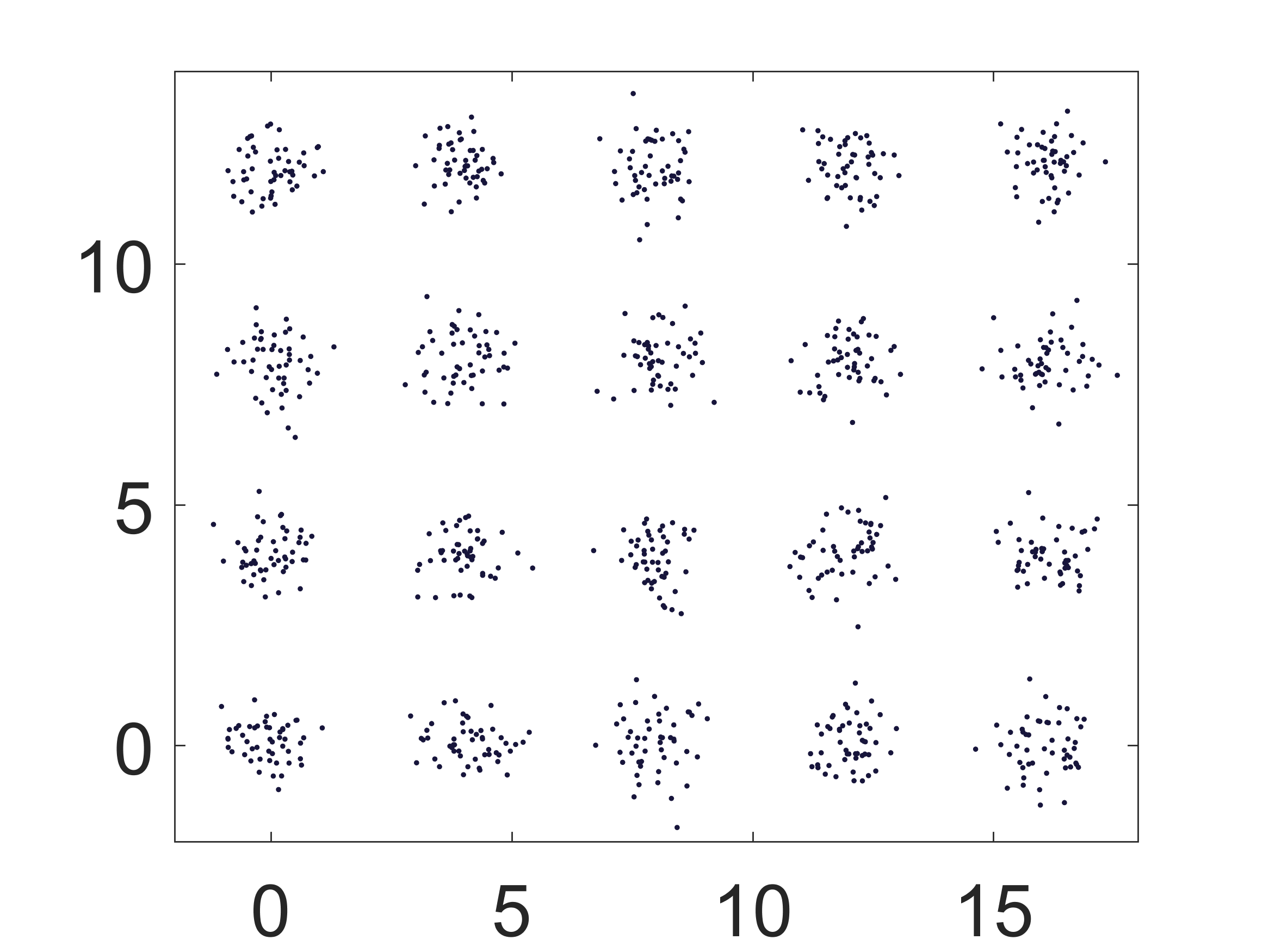}
        \caption{\textsc{Twenty}}
        \label{twenty:data}
    \end{subfigure}
    \begin{subfigure}{0.49\textwidth}
        \centering
        \includegraphics[scale=0.245]{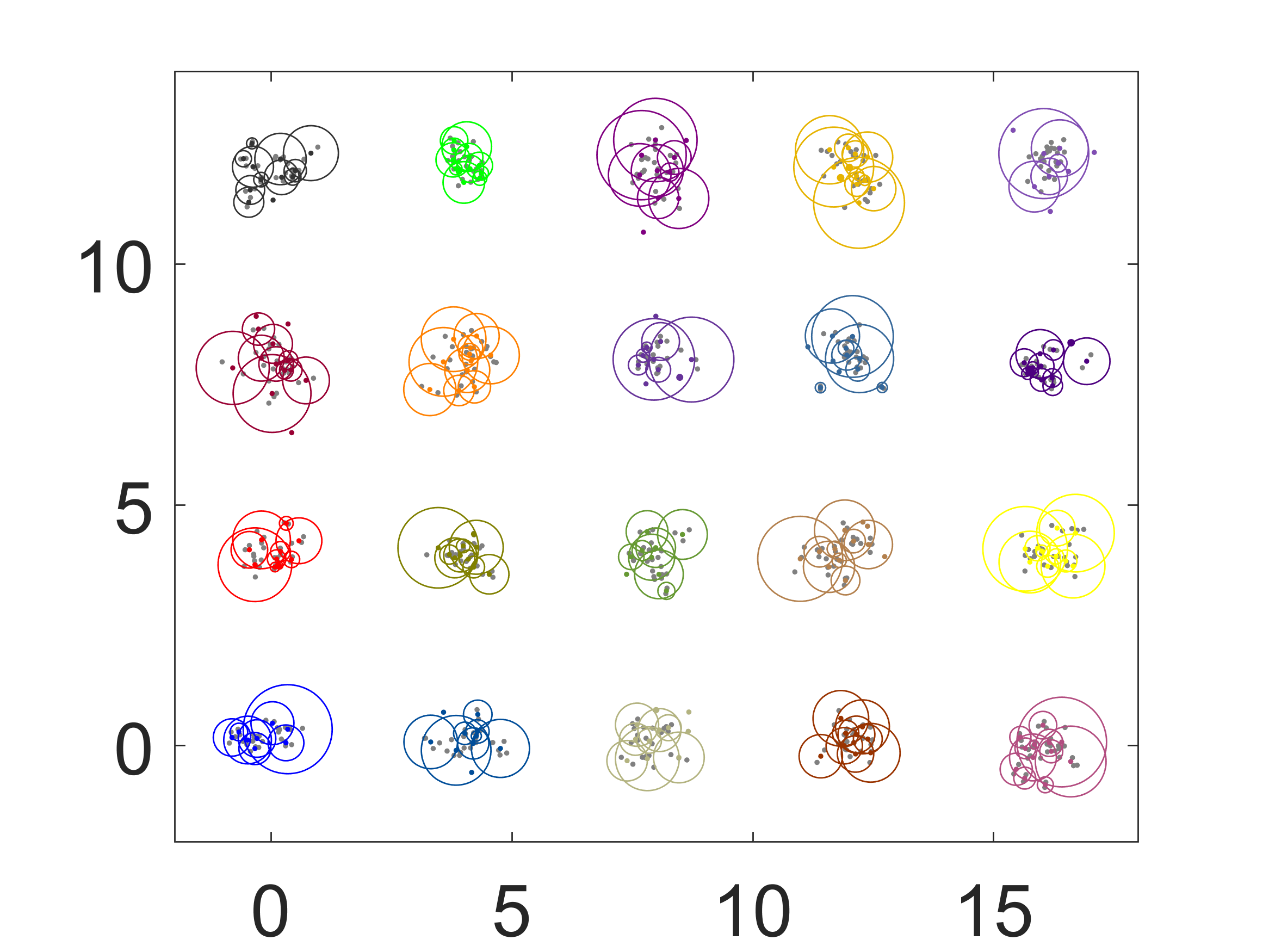}
        \includegraphics[scale=0.245]{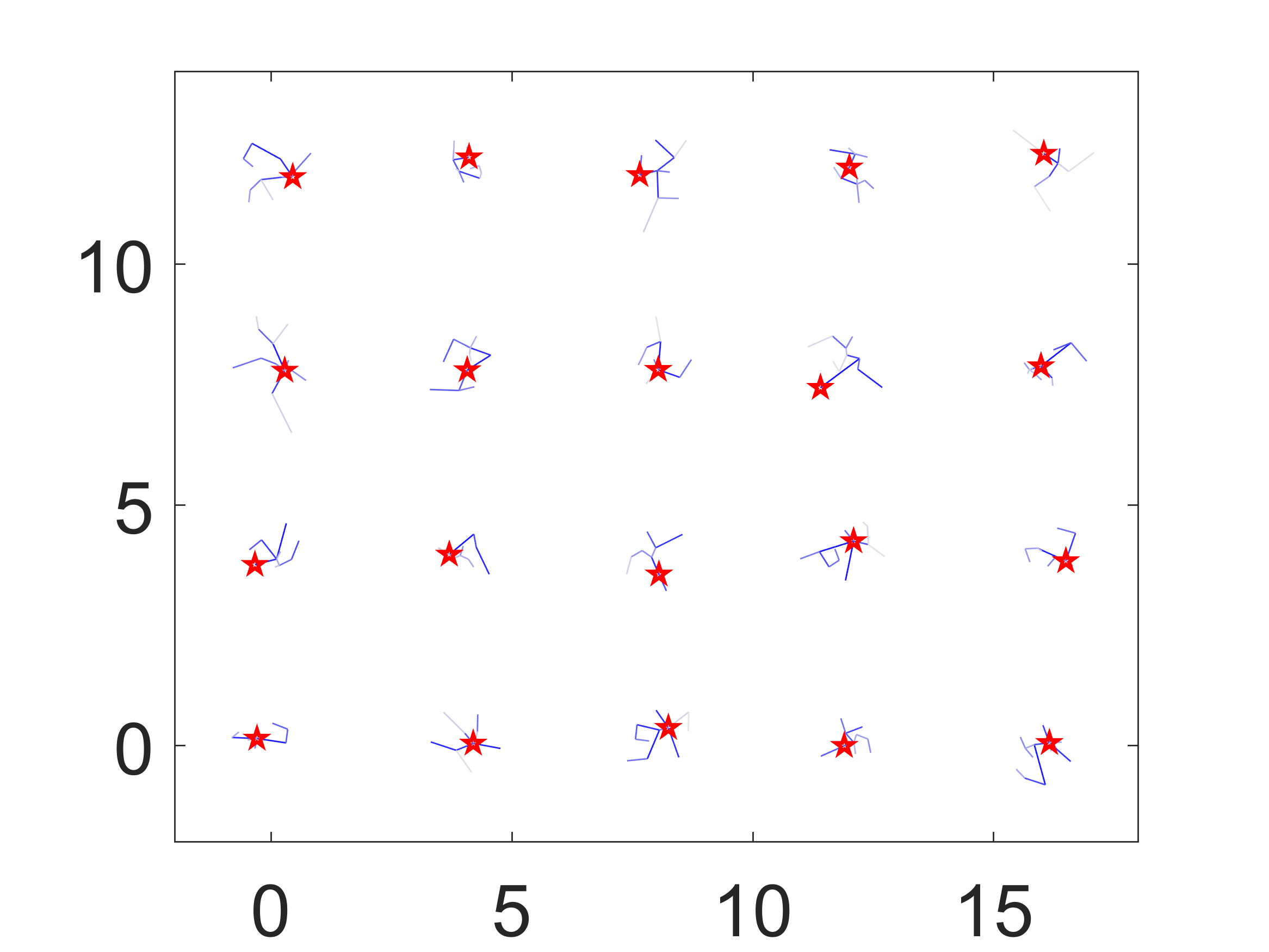}
        \caption{Key granular-balls and skeleton}
        \label{twenty:samples_and_skeleton}
    \end{subfigure}
    \begin{subfigure}{0.245\textwidth}
        \centering
        \includegraphics[scale=0.245]{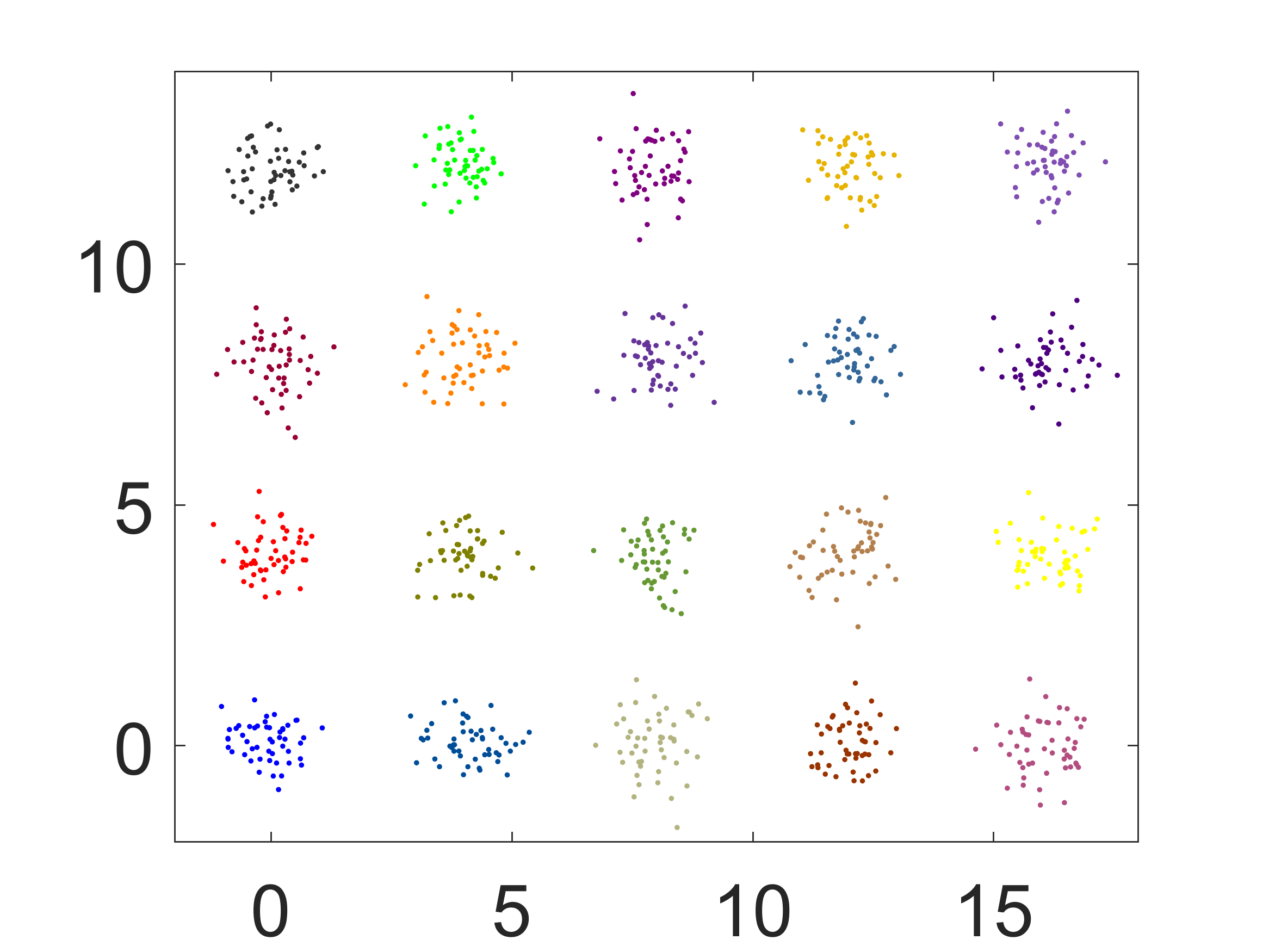}
        \caption{Result}
        \label{twenty:our_label}
    \end{subfigure}
    
    \caption{2D example on \textsc{Twenty}.
    }
    \label{fig:twenty} 
    \vspace{-9pt} 
\end{figure*}

\begin{figure*}[tbh]
    \centering
    
    \begin{subfigure}{0.245\textwidth}
        \centering
        \includegraphics[scale=0.245]{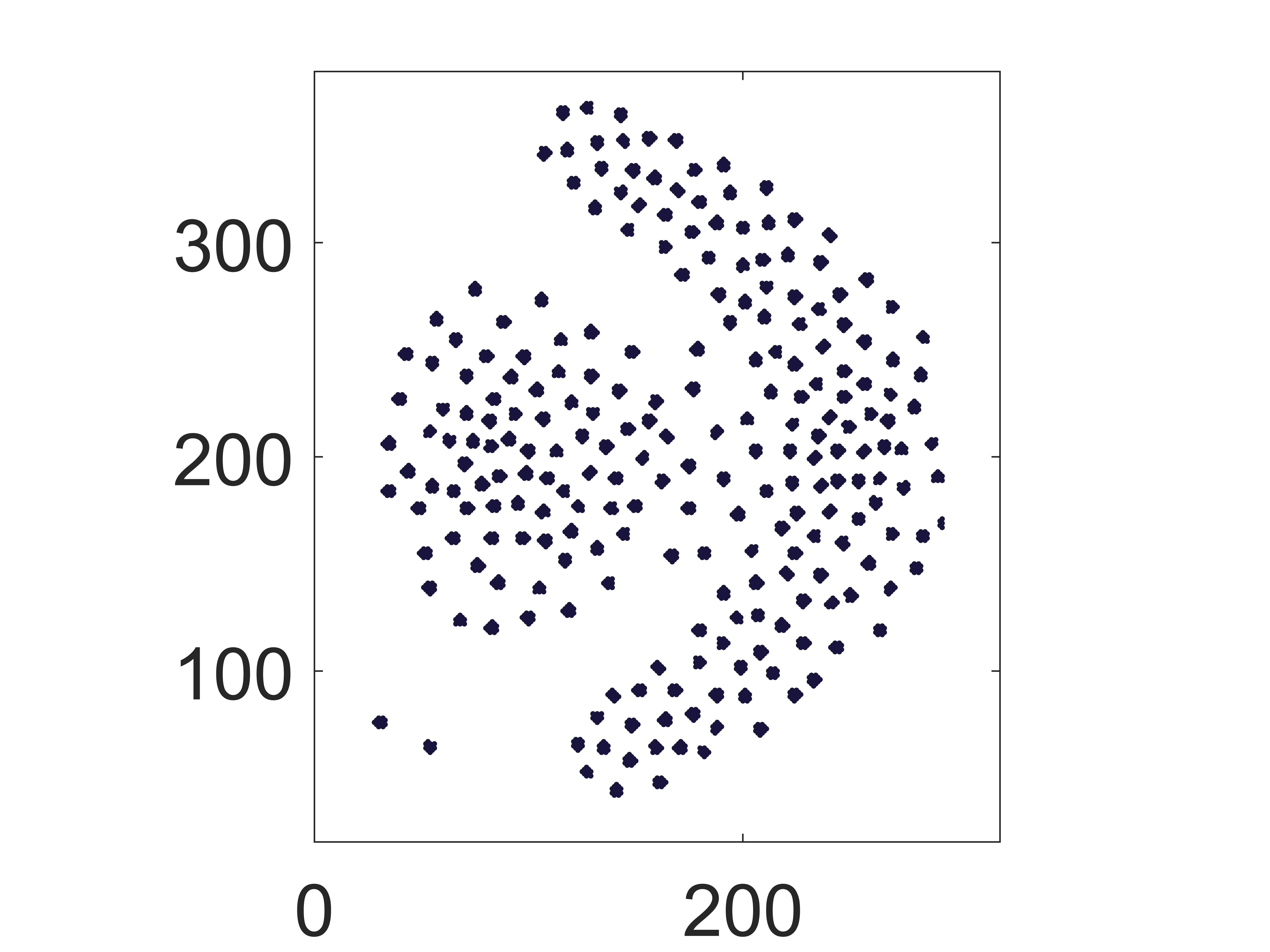}
        \caption{\textsc{SYN3}}
        \label{SYN3:data}
    \end{subfigure}
    \begin{subfigure}{0.49\textwidth}
        \centering
        \includegraphics[scale=0.245]{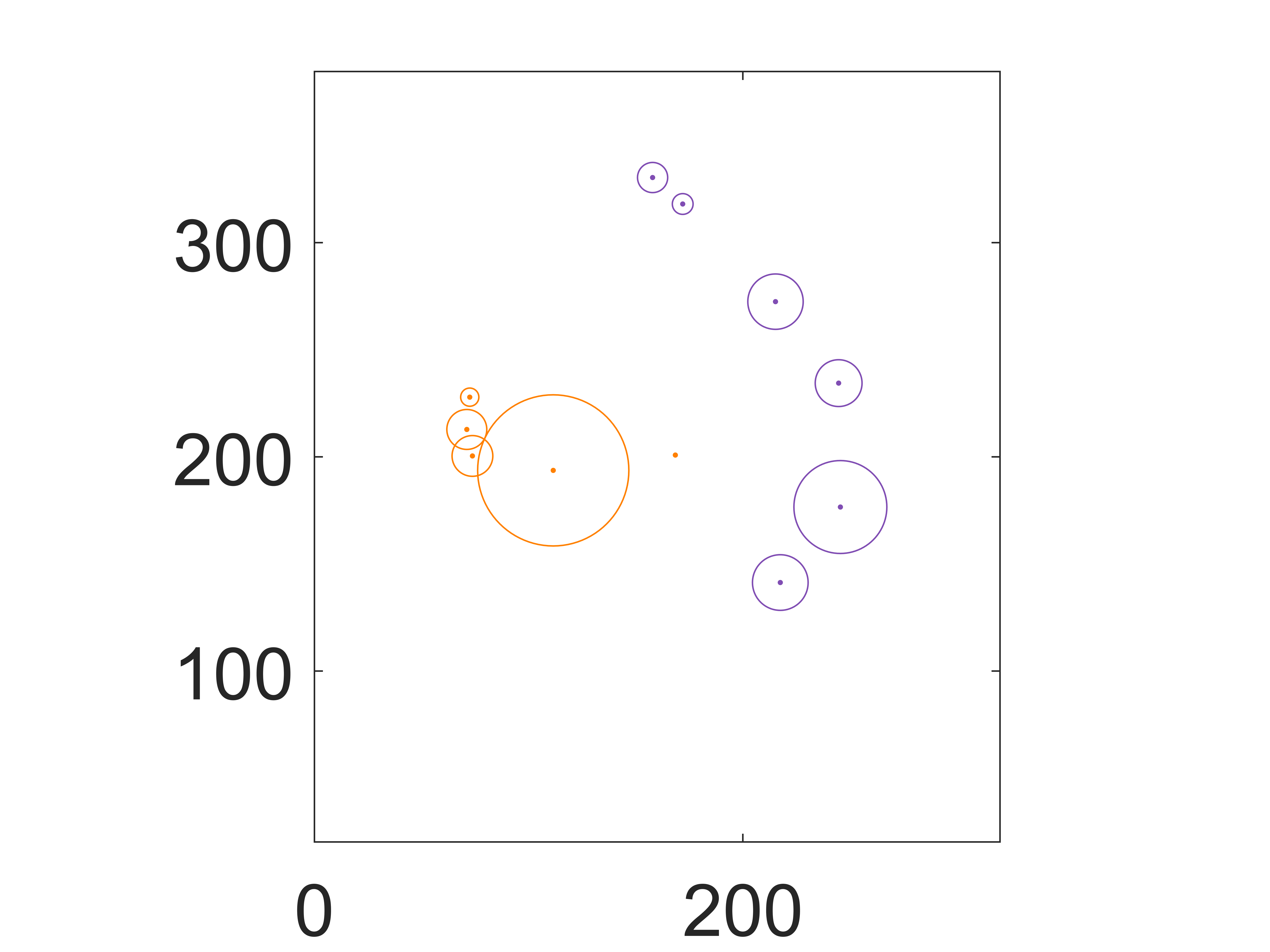} 
        \includegraphics[scale=0.245]{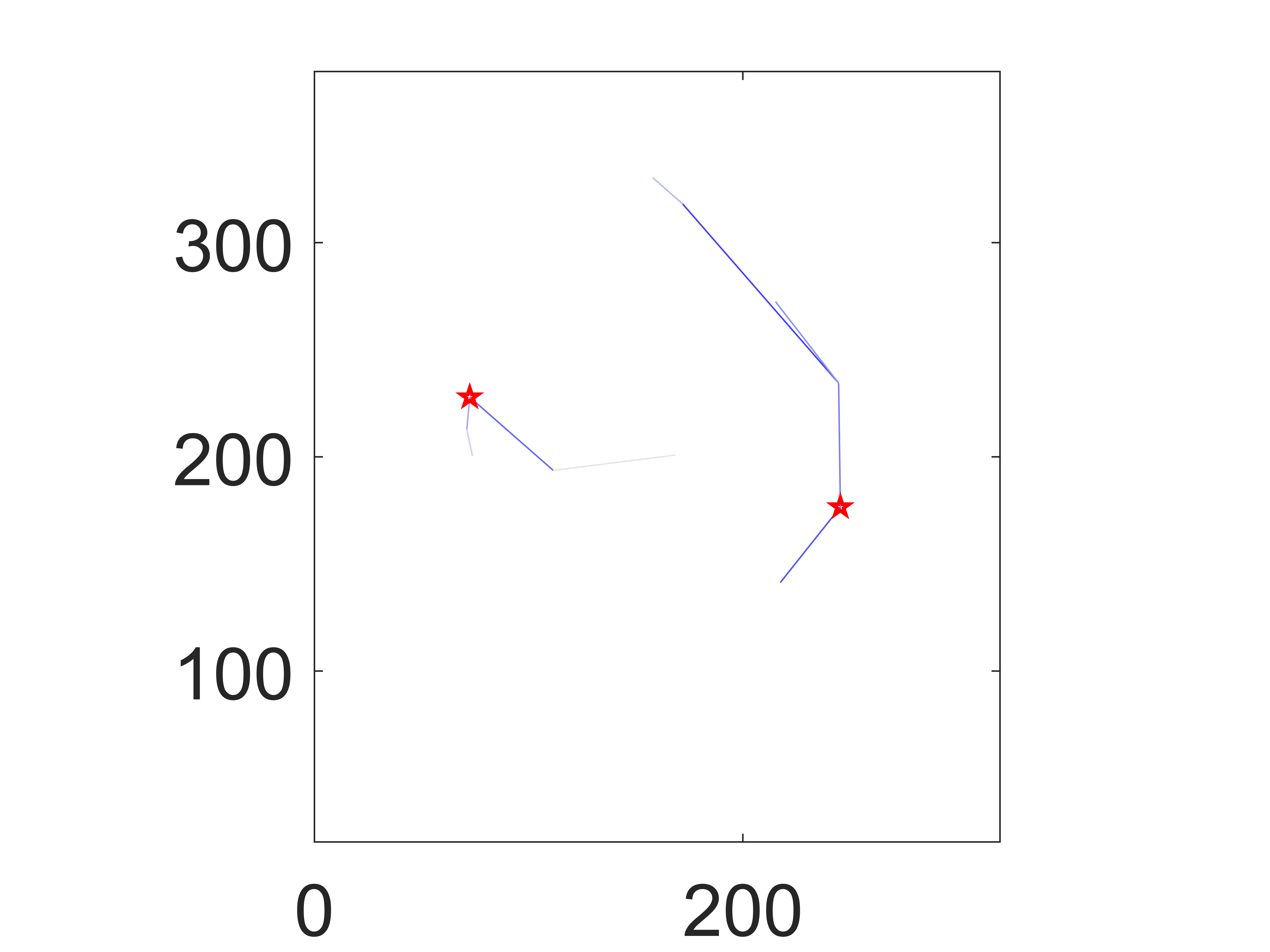}
        \caption{Key granular-balls and skeleton}
        \label{SYN3:skeleton_and_label}
    \end{subfigure}
    \begin{subfigure}{0.245\textwidth}
        \centering
        \includegraphics[scale=0.245]{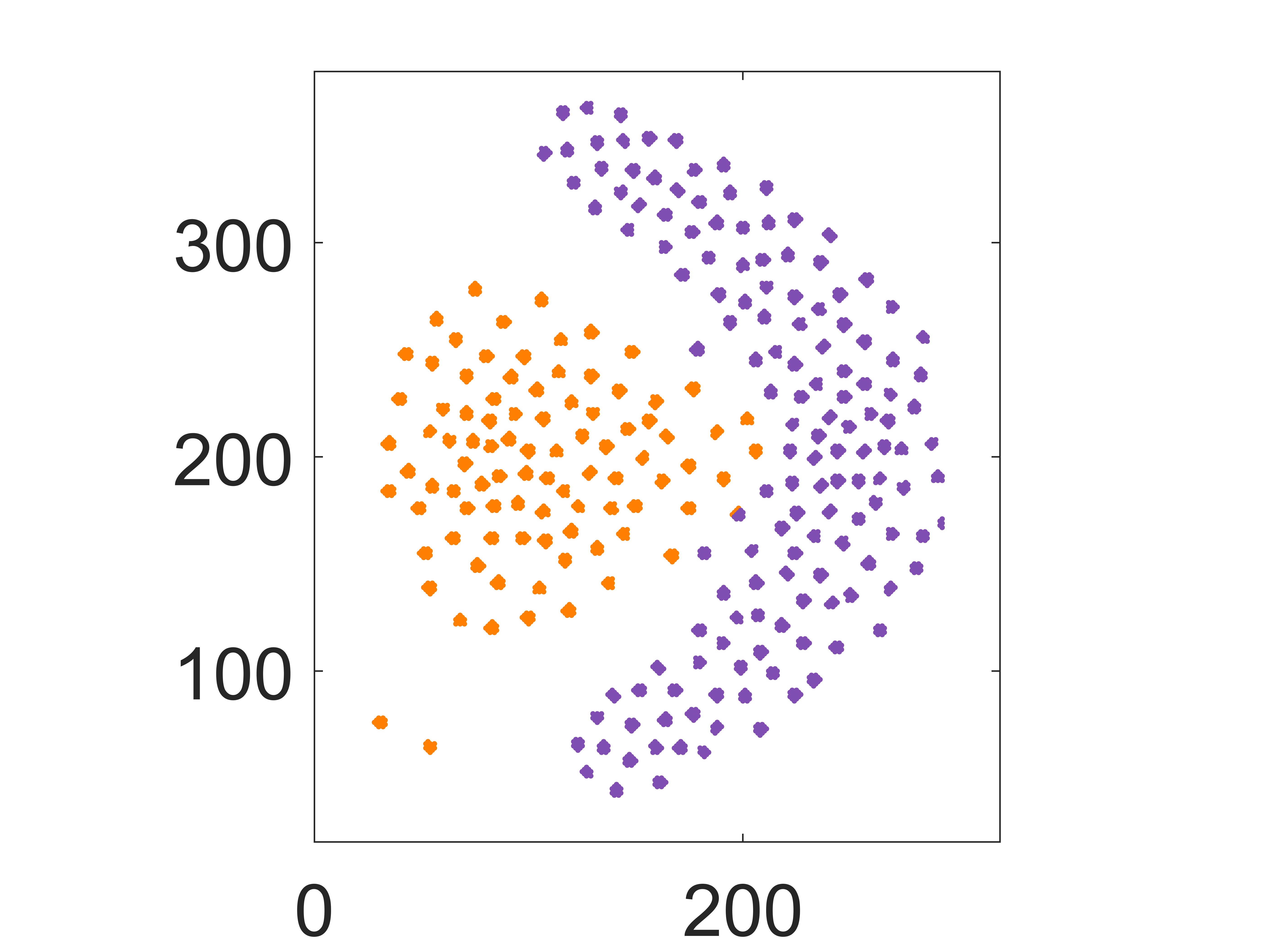}
        \caption{Result}
        \label{SYN3:DP_label}
    \end{subfigure}
    
    \caption{ 2D example on \textsc{SYN3}.
    }
    \label{fig:SYN3}
    \vspace{-9pt}
\end{figure*}

\begin{figure*}[tbh]
    \centering
    
    \begin{subfigure}{0.245\textwidth}
        \centering
        \includegraphics[scale=0.245]{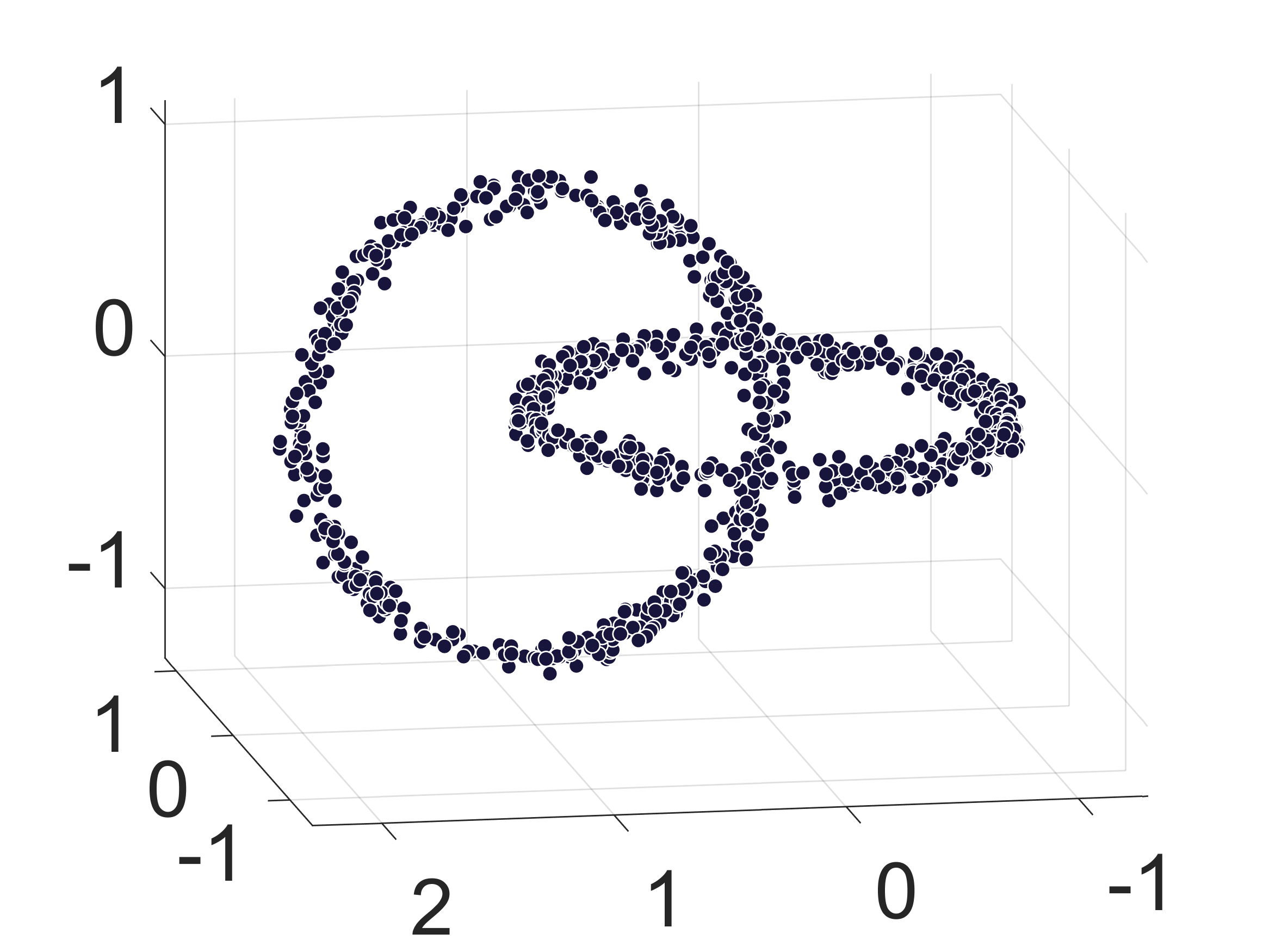}
        \caption{\textsc{Chainlink}}
        \label{Chainlink:data}
    \end{subfigure}
    \begin{subfigure}{0.49\textwidth}
        \centering
        \includegraphics[scale=0.245]{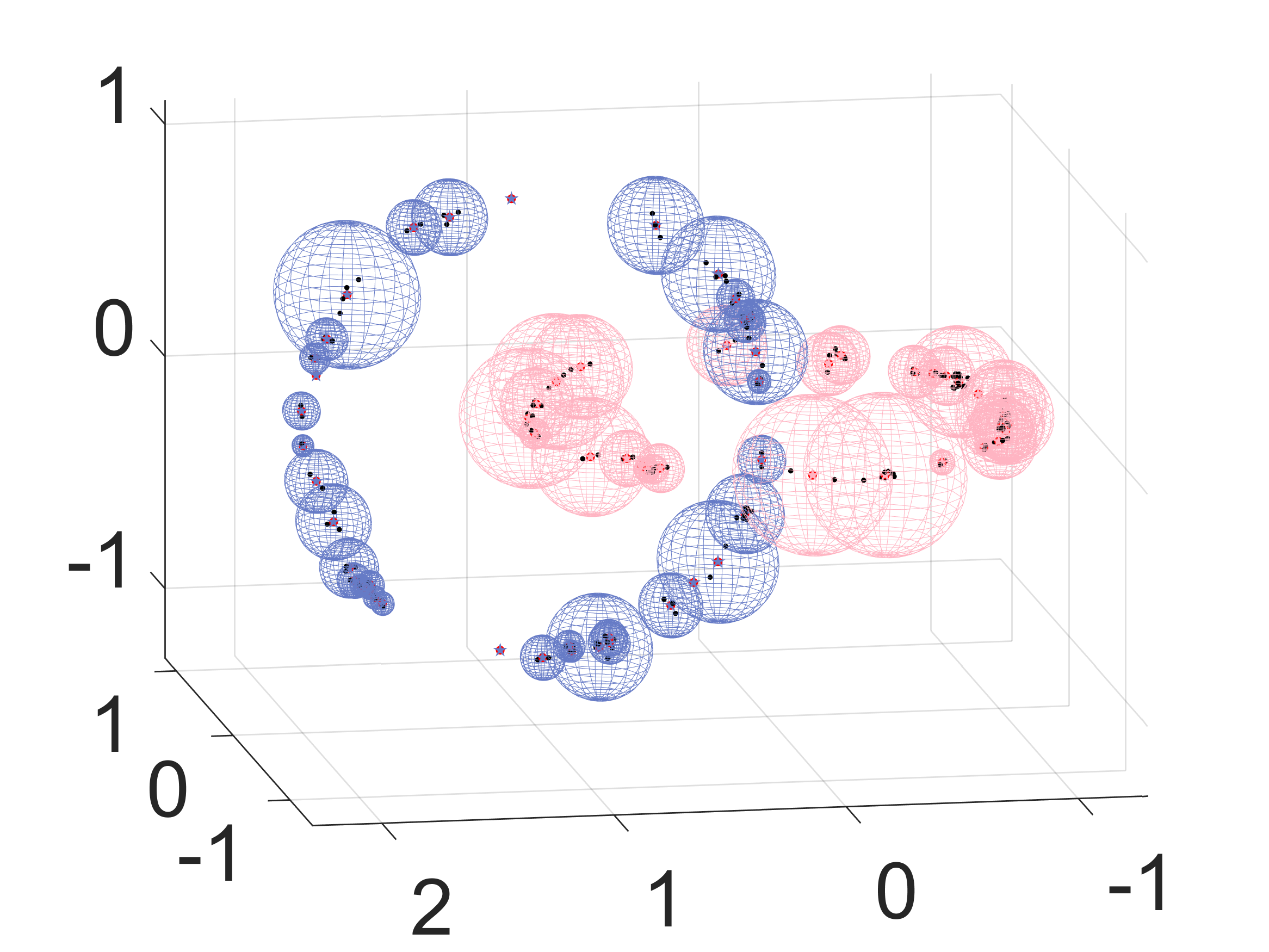}
        \includegraphics[scale=0.245]{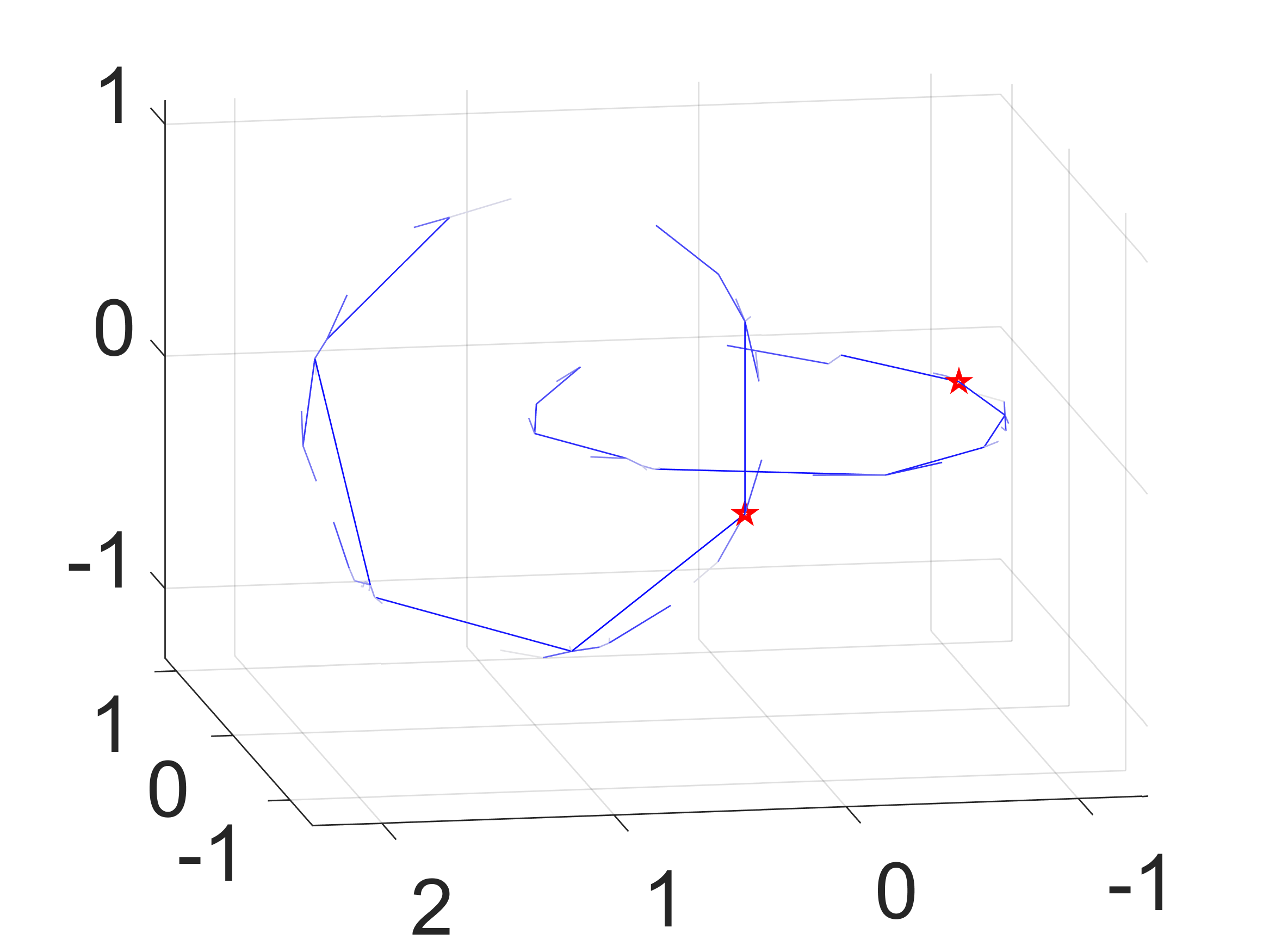}
        \caption{Key granular-balls and skeleton}
        \label{Chainlink:KEYBALLS_and_Skeleton}
    \end{subfigure}
    \begin{subfigure}{0.245\textwidth}
        \centering
        \includegraphics[scale=0.245]{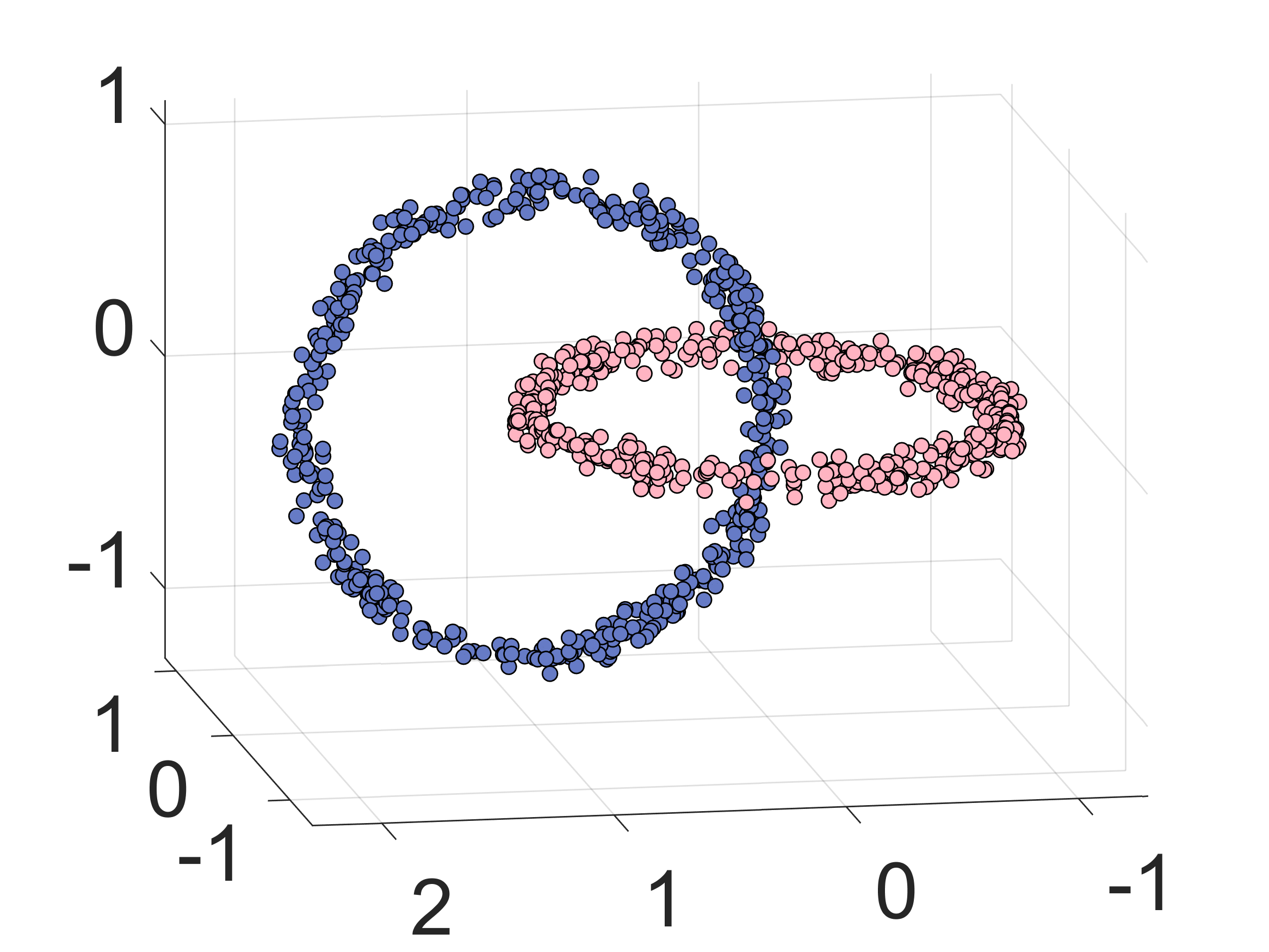}
        \caption{Result}
        \label{Chainlink:result}
    \end{subfigure}
    
    \caption{3D example on \textsc{ChainLink}.
    }
    \label{fig:Chainlink}
    \vspace{-9pt}
\end{figure*}

\begin{itemize}
    \item \textbf{Number of sample sets \boldmath$s$\unboldmath}: Choose $s$ in the range $10 \leq s \leq 50$ to achieve a good trade-off between sampling diversity and computational cost. A setting of $s = 30$ is often sufficient in practice. 
    
    \item \textbf{Sampling proportion \boldmath$\alpha$\unboldmath}: Set $\alpha = \frac{1}{\sqrt{n}}$, where $n$ is the number of data points. This allows the method to maintain scalability while ensuring sufficient sampling coverage. 
    
    \item \textbf{Number of granular-balls \boldmath$M$\unboldmath}: Set $M = 10k$, where $k$ is the number of clusters. This provides an adequate number of granular-balls to capture the internal structure of the data. 
    In general, $k < M \ll n$.
\end{itemize}

\subsection{AGBSK: Simplified GBSK with Default Parameters}\label{subsec:AGBSK_param}
To further improve accessibility and reduce the burden of parameter configuration, we provide a simplified variant of GBSK, named \textbf{AGBSK} (Adaptive GBSK). AGBSK is designed for users who prioritize ease of use or operate in environments where tuning is impractical. 

AGBSK adopts the recommended defaults discussed above and requires only the cluster number $k$ as input. The full configuration is:
\[
[s = 30, \alpha = \frac{1}{\sqrt{n}}, M = 10k, k].
\]

By automating parameter selection, AGBSK balances usability and performance. It leverages sampling for scalability and granular-ball techniques for accurate clustering, making it ideal for large-scale applications.

\subsection{Complexity Analysis}
\label{subsec:complxity analysis}
Algorithm \ref{alg:generateBalls}-\ref{alg:GBSK} detail the GBSK method, which follows a local-global processing structure. Its time and space complexities are presented below.

\begin{table*}
    \centering
    \begin{threeparttable}
        \captionof{table}{Clustering quality of different algorithms on small-scale synthetic and real datasets.} 
        \label{tab: clustering quality}
        \begin{tabularx}{\textwidth}{l cccc cccc cccc}
        \toprule
        Dataset &  \multirow{2}{*}[-0.6em]{\shortstack{Parameters \\ $[s,\alpha,M,k]$}} & \multicolumn{3}{c}{\textsc{S3}} & \multirow{2}{*}[-0.6em]{\shortstack{Parameters \\ $[s,\alpha,M,k]$}} & \multicolumn{3}{c}{\textsc{EngyTime}} & \multirow{2}{*}[-0.6em]{\shortstack{Parameters \\ $[s,\alpha,M,k]$}} & \multicolumn{3}{c}{\textsc{Twenty}}\\
        \cmidrule(l{0.65em}r{1.0em}){3-5}
        \cmidrule(l{0.65em}r{1.0em}){7-9}
        \cmidrule(l{0.65em}r{1.0em}){11-13}
        Method &   & ACC & ARI & AMI &   & ACC & ARI & AMI &    & ACC & ARI & AMI \\
        \midrule
        \multirow{4}{*}{GBSK} & \textbf{[35,0.05,40,15]} & \textbf{0.938} & \textbf{0.878} & \textbf{0.900} & \textbf{[10,-,-,2]} & \textbf{0.966} & \textbf{0.868} & \textbf{0.787} & [20,0.1,30,20] & 1.000 & 1.000 & 1.000\\
        
        & [20,-,-,15] & 0.900 & 0.814 & 0.870 & [20,-,-,2] & 0.964 & 0.861 & 0.776 & [20,0.1,40,20] & 1.000 & 1.000 & 1.000\\
        
        & [30,-,-,15] & 0.891 & 0.793 & 0.859 & [30,-,-,2] & 0.957 & 0.834 & 0.762 & [30,0.1,80,20] & 1.000 & 1.000 & 1.000\\
        
        & [40,-,-,15] & 0.917 & 0.840 & 0.877 & [40,-,-,2] & 0.965 & 0.864 & 0.789 & [20,0.25,-,20] & 1.000 & 1.000 & 1.000\\
        
        \hline
        DPeak \cite{rodriguez2014clustering}& & 0.855 & 0.725 & 0.795 &   & 0.965 & 0.863 & 0.784 &   & 1.000 & 1.000 & 1.000\\
        GDPC \cite{xu2018improved} & & 0.852 & 0.720 & 0.793 &   & 0.959 & 0.857 & 0.781 &   & 1.000 & 1.000 & 1.000\\
        CDPC \cite{xu2018improved} & & 0.850 & 0.715 & 0.790 &   & 0.965 & 0.863 & 0.784 &   & 1.000 & 1.000 & 1.000\\
        FSDPC \cite{xu2021fast}& & 0.855 & 0.725 & 0.795 &   & 0.965 & 0.863 & 0.784 &   & 1.000 & 1.000 & 1.000\\
        FHC-LDP \cite{guan2021fast}& & 0.852 & 0.721 & 0.794  &  & 0.628 & 0.066 & 0.178 &   & 1.000 & 1.000 & 1.000\\
        SDPC \cite{ding2023sampling}& & 0.846 & 0.710 & 0.788  &   & \textbf{0.966} & \textbf{0.868} & \textbf{0.787} &   & 1.000 & 1.000 & 1.000\\
        \hline
        \\
        Dataset & \multirow{2}{*}[-0.6em]{\shortstack{Parameters \\ $[s,\alpha,M,k]$}} & \multicolumn{3}{c}{\textsc{Segmentation}} & \multirow{2}{*}[-0.6em]{\shortstack{Parameters \\ $[s,\alpha,M,k]$}} & \multicolumn{3}{c}{\textsc{Waveform}} & \multirow{2}{*}[-0.6em]{\shortstack{Parameters \\ $[s,\alpha,M,k]$}} & \multicolumn{3}{c}{\textsc{Pendigits}}\\
        \cmidrule(l{0.65em}r{1.0em}){3-5}
        \cmidrule(l{0.65em}r{1.0em}){7-9}
        \cmidrule(l{0.65em}r{1.0em}){11-13}
         Method &  & ACC & ARI & AMI &   & ACC & ARI & AMI &    & ACC & ARI & AMI \\
        \midrule
        \multirow{4}{*}{GBSK} & [30,-,-,7] & 0.568 & 0.359 & 0.554 
        & [10,-,-,3] & 0.646 & 0.336 & 0.405 
        & [10,-,-,10] & 0.765 & 0.580 &  0.703\\
        
        & [50,-,-,7] & 0.600 & 0.475 & 0.603 
        & [20,-,-,3] & 0.643 & 0.334 & 0.408 
        & [20,-,-,10] & 0.768 & 0.602 & 0.713\\
        
        & [70,-,-,7] & 0.628 & 0.445 & 0.577
        & [30,-,-,3] & 0.683 & 0.364 & 0.421 
        & \textbf{[30,-,-,10]} & \textbf{0.793} & 0.641 & \textbf{0.738}\\
        
        & [100,-,-,7] & 0.630 & 0.443 & 0.579
        & \textbf{[30,0.05,40,3]} & \textbf{0.723} & \textbf{0.404} & \textbf{0.452}
        & [20,0.02,40,10] & 0.786 & \textbf{0.645} & 0.733\\
        \hline
        DPeak & & 0.598 & 0.407 & 0.564 &   & 0.542 & 0.255 & 0.369 &   & 0.650 & 0.537 & 0.683\\
        GDPC & & 0.430 & 0.293 & 0.484 &   & 0.521 & 0.250 & 0.359 &   & 0.641 & 0.523 & 0.654\\
        CDPC & & 0.546 & 0.367 & 0.486 &   & 0.503 & 0.245 & 0.352 &   & 0.648 & 0.533 & 0.667\\
        FSDPC & & 0.597 & 0.407 & 0.564 &   & 0.541 & 0.255 & 0.369 &   & 0.650 & 0.536 & 0.682\\
        FHC-LDP & & \textbf{0.684} & \textbf{0.543} & \textbf{0.656} &   & 0.616 & 0.258 & 0.331 &   & 0.715 & 0.604 & 0.737\\
        SDPC & & 0.660 & 0.385 & 0.562 &   & 0.597 & 0.281 & 0.382 &   & 0.649 & 0.478 & 0.614\\
        \bottomrule
        \end{tabularx}
        \begin{tablenotes}
            \item The results of competitors in this table are all excerpted from \cite{ding2023sampling} and are independent from hardware devices. A ``-'' in Parameters indicates AGBSK default settings. 
        \end{tablenotes}
    \end{threeparttable}
\end{table*}

Algorithm \ref{alg:generateBalls} employs k-means++ (k=2) to generate balls, with time complexity $O(n)$; 
Algorithm \ref{alg:identifyPeakBalls} runs in $O(M^2)$ expected time, where $M$ is the number of granular-balls; 
Algorithm \ref{alg:constructForest} performs in $O(W^2)$ time, with $W$ as the number of key balls; 
Algorithm \ref{alg:GBSK} consists of five steps: 

Step 1: Randomly sample $s$ sets, each of size $n \times \alpha$. Thus, the time complexity is $O(s \times n \times \alpha)$;

Step 2: For each $P^{(i)}$, the time of generating representative balls is $O(n \times \alpha + M^2)$. Since $n \times \alpha \gg M^2$, it runs in $O(s \times \alpha \times  n)$;

Step 3: The number of representative balls is approximately $s \times k$. Using Algorithm \ref{alg:generateBalls}, the time is $O(s \times k)$;

Step 4: The same as Algorithm \ref{alg:constructForest}: $O(W^2)$ ;

Step 5: Final step runs in $O(W \times n)$. 

Comprehensively, the overall time complexity of GBSK is $O(s \times \alpha \times n + s \times k + W^2 + W \times n) = O(s \times \alpha \times n + W \times n)$. AGBSK runs in $O(s \times \sqrt{n} + W \times n) = O(n)$, where $W < k \times s < M \times s$ and $W \ll M \times s \ll n$. 

The overall space complexity of GBSK is $O(n + \alpha \times n + s \times k + W + n)=O(n)$.

\section{EXPERIMENTS}\label{sec:experiments}
This section aims to demonstrate the scalability, efficiency, effectiveness, and usability of the proposed method on complex, large-scale, high-dimensional datasets, highlighting its advantages over traditional clustering methods.

\subsection{Experimental Settings and Methods}

\textit{Setup}: GBSK is implemented in MATLAB 2021 and runs on standard hardware with Windows 10 (64-bit), Intel(R) Core(TM) i7-10700 CPU @ 2.90GHz, 128GB RAM. 

\textit{Datasets}: There are nine synthetic datasets (\textsc{Chainlink}, \textsc{EngyTime}, \textsc{SYN1}\cite{chen2020fast}, \textsc{SYN2}\cite{chen2020fast}, \textsc{SYN3}\cite{chen2020fast}, \textsc{S3}, \textsc{Twenty}, \textsc{3M2D5}\cite{zhao2019stratified}, \textsc{AGC100M}) and ten real-world datasets (\textsc{Pendigits}, \textsc{DryBean}, \textsc{Waveform}, \textsc{Segmentation}, \textsc{CIFAR-10}, \textsc{MoCap}, \textsc{CoverType}, \textsc{N-BaIoT}, \textsc{MNIST}, \textsc{MNIST8M}). Appendix~F Table~V shows the details. 

\textit{Competitors}: Since our method uses Algorithm~\ref{alg:identifyPeakBalls}, similar to density peaks, in Step 2 and Step 4, it is natural to benchmark it against notable density peak-based clustering algorithms. This includes DPeak \cite{rodriguez2014clustering}, GDPC \cite{xu2018improved}, CDPC \cite{xu2018improved}, FSDPC \cite{xu2021fast}, FHC-LDP \cite{guan2021fast}, SDPC \cite{ding2023sampling}, FastDPeak \cite{chen2020fast}, Fast LDP-MST \cite{qiu2022fast}, GB-DP \cite{cheng2023fast}. Nevertheless, k-means++ is included as a widely recognized and commonly used baseline.

\textit{Evaluation metrics}: accuracy (ACC), adjusted rand index (ARI), adjusted mutual information (AMI). 

\subsection{Clustering Quality on Benchmark Datasets}
To evaluate the correctness of the proposed algorithm, we conducted two types of experiments. The first type focuses on  visualizing clustering outcomes on 2D and 3D datasets. The second compares the results of GBSK against competitors on synthetic and real-world datasets, using the three metrics. 

\textit{On 2D and 3D datasets}: We evaluated the performance of our proposed GBSK algorithm on four datasets: \textsc{SYN1}, \textsc{SYN3}, \textsc{Twenty} and \textsc{Chainlink}. The results are shown in Fig.~\ref{fig:SYN1}-\ref{fig:Chainlink}. GBSK demonstrates strong performance, particularly on the complex structure of \textsc{Chainlink}, where it outperforms both DPeak and SDPC. 

\textit{On small-scale synthetic and real datasets}: Experiments were conducted to compare the ACC, ARI, and AMI metrics on three synthetic datasets: \textsc{S3}, \textsc{EngyTime}, and \textsc{Twenty}, as well as three real datasets: \textsc{Segmentation}, \textsc{Waveform}, and \textsc{Pendigits}. 

\begin{table*}[th]
    \centering
    \renewcommand{\arraystretch}{1.15}
    \setlength{\tabcolsep}{2.3pt}
    \caption{Comparisons on large-scale datasets.}
    \label{tab:compare_GBSK_large}
    \begin{tabular}{|c|ccccccc|ccccccc|ccccccc|}
    \hline
    \multirow{3}{*}{Dataset} & \multicolumn{7}{c|}{Parameters} & \multicolumn{7}{c|}{ACC} & \multicolumn{7}{c|}{Runtime (unit: seconds)}\\
    \cline{2-22}
    & AGBSK & GBSK & FD & FL & FM & GD & KM+ & \multirow{2}{*}{AGBSK} & \multirow{2}{*}{GBSK} & \multirow{2}{*}{FD} & \multirow{2}{*}{FL} & \multirow{2}{*}{FM} & \multirow{2}{*}{GD} & \multirow{2}{*}{KM+} & \multirow{2}{*}{AGBSK} & \multirow{2}{*}{GBSK} & \multirow{2}{*}{FD} & \multirow{2}{*}{FL} & \multirow{2}{*}{FM} & \multirow{2}{*}{GD} & \multirow{2}{*}{KM+} \\
    & $k$ & $[s,\alpha,M,k]$ & $K$ & $K$ & $K$ & $k$ & $k$ &  &  &  &  &  &  &  &  &  &  &  &  &  &  \\
    \hline
     
    \multirow{4}{*}{\parbox{1.4cm}{\centering \textsc{Pendigits}\\n=1E4\\d=16}}
    & \multirow{4}{*}{10} & [10,-,-,10] & 5 & 20 & 20 & \multirow{4}{*}{10} & \multirow{4}{*}{10}
    & \multirow{4}{*}{\underline{0.80}} & 0.79 & 0.76  & 0.58  & 0.69  & \multirow{4}{*}{0.65} & \multirow{4}{*}{0.74}
    & \multirow{4}{*}{1.4} & 0.6 & 0.5 & 5.1 & 0.5 & \multirow{4}{*}{1.7} & \multirow{4}{*}{\textbf{0.05}}\\
    && [-,1E-3,-,10] & 10 & 25 & 25 & &
    && 0.67 & 0.77  & 0.69  & 0.68  &       &
    && \underline{0.2} & 0.5 & 4.5 & 0.6 &&\\
    && [-,-,10,10] & 15    & 30    & 45    & &
    && 0.76 & 0.76  & 0.69  & 0.68  &       &
    && 0.9 & 0.7 & 4.9 & 1.3 &&\\
    && [50,-,50,10] & 20    & 35    & 50    & &
    && \textbf{0.81} & 0.67  & 0.65  & 0.68  &       & 
    && 2.6 & 0.7 & 4.8 & 1.6 &&\\
    \hline

    \multirow{4}{*}{\parbox{1.4cm}{\centering \textsc{DryBean}\\n=1E4\\d=16}}  
    & \multirow{4}{*}{7} & [10,1E-3,7,7] & 5 & 20    & 5     & \multirow{4}{*}{7} & \multirow{4}{*}{7}
    & \multirow{4}{*}{0.60} & \underline{0.61} & 0.44  & 0.57  & 0.32  & \multirow{4}{*}{0.34} & \multirow{4}{*}{0.59} 
    & \multirow{4}{*}{1.9} & \underline{0.1} & 0.5 & 9.9 & 0.2 & \multirow{4}{*}{1.7} & \multirow{4}{*}{\textbf{0.08}} \\
    && [10,-,-,7] & 10    & 25    & 10    &       &
    && 0.60 & 0.48  & 0.54  & 0.36  &       & 
    && 0.7 & 0.5 & 9.2 & 0.3 &&\\
    && [-,1E-3,-,7] & 15    & 30    & 15    &       &
    && \underline{0.61} & 0.54 & \textbf{0.63} & 0.30 & & 
    && 0.2 & 0.5 & 10.4 & 0.6 &&\\
    && [-,-,7,7] & 20    & 35    & 20    &       &
    && 0.60 & 0.43  & 0.60  & 0.30  &&
    && 0.4 & 0.5 & 8.3 & 0.6  &&\\
    \hline
    
    \multirow{4}{*}{\parbox{1.4cm}{\centering \textsc{MoCap}\\n=8E4\\d=36}}
    & \multirow{4}{*}{5} & [10,-,25,5] & 5 & 20 & 15 & \multirow{4}{*}{5} & \multirow{4}{*}{5}
    & \multirow{4}{*}{\underline{0.67}} & \textbf{0.68} & 0.37 & 0.21 & 0.21 & \multirow{4}{*}{0.26} & \multirow{4}{*}{0.54}
    & \multirow{4}{*}{2.0} & \textbf{0.4} & 90 & 26 & 5.4 & \multirow{4}{*}{5.7} & \multirow{4}{*}{\underline{0.7}}\\
    && [10,-,-,5] & 10 & 25 & 20 && 
    && 0.64 & 0.26 & 0.21 & 0.21 &&
    && \underline{0.7} & 82 & 24 & 6.3 &&\\
    && [-,1E-2,-,5] & 15 & 30 & 25 && 
    && \textbf{0.68} & 0.41 & 0.21 & 0.22 &&
    && 2.1 & 84 & 24 & 7.6 &&\\
    && [-,-,25,5] & 20 & 35 & 30 && 
    && 0.66 & 0.45 & 0.21 & 0.26 &&
    && 1.0 & 88 & 25 & 8.8 &&\\
    \hline
         
    \multirow{4}{*}{\parbox{1.4cm}{\centering \textsc{CoverType}\\n=6E5\\d=54}} 
    & \multirow{4}{*}{7} & [-,1E-4,35,7] & 5 & 20    & 10    & \multirow{4}{*}{7} & \multirow{4}{*}{7} 
    & \multirow{4}{*}{0.40} & 0.41 & 0.41 & $\infty$ & \textbf{0.49} & \multirow{4}{*}{0.39} & \multirow{4}{*}{0.25} 
    & \multirow{4}{*}{4.0} & \textbf{1.3} & 1251 & $\infty$ & 68 & \multirow{4}{*}{37} & \multirow{4}{*}{14} \\
    && [50,-,-,7] & 10    & 25    & 15    &       &
    && 0.42 & 0.34 & $\infty$ & \textbf{0.49} &&
    && 7.0 & 327    & $\infty$  & 88 &       &  \\
    && [-,1E-4,-,7] & 15    & 30    & 20    &       &
    && 0.40 & 0.26  & $\infty$  & \underline{0.43}  &       &
    && \underline{1.4} & 171   & $\infty$  & 85 &       &  \\
    && [-,-,35,7] & 20   & 35    & 25    &       &
    && 0.39 & 0.28 & $\infty$ & \underline{0.43} &&
    && 2.6 & 127   & $\infty$   & 90  &       &  \\
    \hline

    \multirow{4}{*}{\parbox{1.4cm}{\centering \textsc{3M2D5 }\\n=3E6\\d=2}} 
    & \multirow{4}{*}{5} & [10,1E-5,-,5] & 5 & 20    & 15   & \multirow{4}{*}{5} & \multirow{4}{*}{5} 
    & \multirow{4}{*}{\underline{0.99}} & \underline{0.99} & $\infty$  & $\infty$   & \textbf{1.00}  & \multirow{4}{*}{0.74} & \multirow{4}{*}{\underline{0.99}} 
    & \multirow{4}{*}{2.2} & \textbf{0.3} & $\infty$ & $\infty$ & 20 & \multirow{4}{*}{54} & \multirow{4}{*}{2.6} \\
    && [10,-,-,5] & 10    & 25    & 20    &       &
    && \underline{0.99} & $\infty$ & $\infty$ & \textbf{1.00} &&
    && \underline{0.8} & $\infty$ & $\infty$ & 19 &&\\
    && [-,1E-5,-,5] & 15    & 30    & 25    &       & 
    && \textbf{1.00} & $\infty$ & $\infty$ & \textbf{1.00} &&
    && \underline{0.8} & $\infty$ & $\infty$ & 19 &&\\
    && [-,-,25,5] & 20   & 35   & 30   &       & 
    && \underline{0.99} & $\infty$ & $\infty$ & \textbf{1.00} &&
    && 1.4 & $\infty$ & $\infty$ & 20 &&\\
    \hline
    
    \multirow{4}{*}{\parbox{1.4cm}{\centering \textsc{MNIST}\\n=7E4\\d=784}}
    & \multirow{4}{*}{10} & [20,1E-3,-,10] & 5 & 20 & 15 & \multirow{4}{*}{10} & \multirow{4}{*}{10} 
    & \multirow{4}{*}{0.56} & 0.56 & $\infty$ & $\infty$ & \textbf{0.72} & \multirow{4}{*}{0.11} & \multirow{4}{*}{0.58}
    & \multirow{4}{*}{6.8} & \textbf{1.3} & $\infty$ & $\infty$ & 39 & \multirow{4}{*}{55} & \multirow{4}{*}{22}\\
    && [30,-,-,10] & 10 & 25 & 20 && 
    && 0.59 & $\infty$ & $\infty$ & 0.61 &&
    && 5.7 & $\infty$ & $\infty$ & 45 &&\\
    && [-,1E-3,-10] & 15 & 30 & 25 && 
    && 0.54 & $\infty$ & $\infty$ & 0.63 &&
    && \underline{2.0} & $\infty$ & $\infty$ & 53 &&\\
    && [-,-,100,10] & 20 & 35 & 30 && 
    && 0.62 & $\infty$ & $\infty$ & \underline{0.64} &&
    && 5.6 & $\infty$ & $\infty$ & 61 &&\\
    \hline

    \multirow{4}{*}{\parbox{1.4cm}{\centering \textsc{CIFAR-10}\\n=6E4\\d=3072}}
    & \multirow{4}{*}{10} & [20,1E-3,-,10] & 5 & 20 & 15 & \multirow{4}{*}{10} & \multirow{4}{*}{10}
    & \multirow{4}{*}{\textbf{0.22}} & \underline{0.21} & $\infty$ & $\infty$ & 0.16 & \multirow{4}{*}{0.10} & \multirow{4}{*}{\underline{0.21}}
    & \multirow{4}{*}{9.1} & \textbf{2.4} & $\infty$ & $\infty$ & 166 & \multirow{4}{*}{153} & \multirow{4}{*}{84}\\
    && [20,-,-,10] & 10 & 25 & 20 && 
    && \textbf{0.22} & $\infty$ & $\infty$ & 0.17 &&
    && 5.8 & $\infty$ & $\infty$ & 201 &&\\
    && [-,1E-3,-,10] & 15 & 30 & 25 && 
    && \textbf{0.22} & $\infty$ & $\infty$ & 0.17 &&
    && \underline{2.9} & $\infty$ & $\infty$ & 248 &&\\
    && [-,-,100,10] & 20 & 35 & 30 && 
    && \textbf{0.22} & $\infty$ & $\infty$ & 0.16 &&
    && 9.0 & $\infty$ & $\infty$ & 283 &&\\
    \hline
    
    \multirow{4}{*}{\parbox{1.4cm}{\centering \textsc{MNIST8M}\\n=8E6\\d=784}}
    & \multirow{4}{*}{10} & [-,1E-5-,10] & 5    & 20    & 15   & \multirow{4}{*}{10} & \multirow{4}{*}{10} 
    & \multirow{4}{*}{0.52} & 0.43 & $\infty$ & $\infty$ & $\infty$ & \multirow{4}{*}{$\infty$} & \multirow{4}{*}{$\infty$}
    & \multirow{4}{*}{184} & \textbf{111} & $\infty$ & $\infty$ & $\infty$ & \multirow{4}{*}{$\infty$} & \multirow{4}{*}{$\infty$}\\
    && [25,-,-,10] & 10    & 25    & 20    &       &
    && \underline{0.53} & $\infty$ & $\infty$ & $\infty$ &&
    && \underline{159} & $\infty$ & $\infty$ & $\infty$ &&\\
    && [-,-,130,10] & 15    & 30    & 25    &       & 
    && \underline{0.53} & $\infty$ & $\infty$ & $\infty$ &&
    && 197 & $\infty$ & $\infty$ & $\infty$ &&  \\
    && [-,-,300,10] & 20   & 35   & 30   &       & 
    && \textbf{0.56} & $\infty$ & $\infty$ & $\infty$ &&
    && 229 & $\infty$   & $\infty$   & $\infty$    &       &  \\
    \hline
    
    \multirow{4}{*}{\parbox{1.4cm}{\centering \textsc{AGC100M}\\n=1E8\\d=256}}
    & \multirow{4}{*}{17} & [10,-,-,17] & 5    & 20    & 15   & \multirow{4}{*}{17} & \multirow{4}{*}{17} 
    & \multirow{4}{*}{\underline{0.95}} & \textbf{0.96} & $\infty$ & $\infty$ & $\infty$ & \multirow{4}{*}{$\infty$} & \multirow{4}{*}{$\infty$} 
    & \multirow{4}{*}{491} & \underline{373} & $\infty$  & $\infty$   & $\infty$ & \multirow{4}{*}{$\infty$} & \multirow{4}{*}{$\infty$} \\
    && [20,-,-,17] & 10    & 25    & 20    &       &
    && \underline{0.95} & $\infty$ & $\infty$ & $\infty$ &&
    && 455 & $\infty$  & $\infty$   & $\infty$    &       &  \\
    && [-,1E-6,-,17] & 15    & 30    & 25    &       &
    && \textbf{0.96} & $\infty$ & $\infty$ & $\infty$ &&
    && \textbf{360} & $\infty$ & $\infty$ & $\infty$ &&\\
    && [-,-,17,17] & 20   & 35   & 30   &       &
    && \underline{0.95} & $\infty$ & $\infty$ & $\infty$ &&
    && 380 & $\infty$ & $\infty$ & $\infty$ &&\\
    \hline
    \end{tabular}%
        \begin{tablenotes}
        \item We simply denote FastDPeak \cite{chen2020fast} as FD, FHC-LDP \cite{guan2021fast} as FL, Fast LDP-MST \cite{qiu2022fast} as FM, GB-DP \cite{cheng2023fast} as GD without parameter, and k-means++ as KM+. 
        The values for ACC and Runtime are highlighted in bold (best) and underlined (second-best). 
        ``$\infty$'' indicates that the runtime exceeds 2000 seconds, and the corresponding algorithm is therefore deemed ineffective. 
        A ``-'' in Parameters denotes the default setting for AGBSK. For instance, ``[10,-,-,10]'' for GBSK (first line in Parameters) corresponds to the parameter setting ``$[s=10,\alpha=\frac{1}{\sqrt{n}},M=10k,k=10]$''. 
    \end{tablenotes}
\end{table*}%

\begin{figure}
    \centering
    \includegraphics[width=1\linewidth]{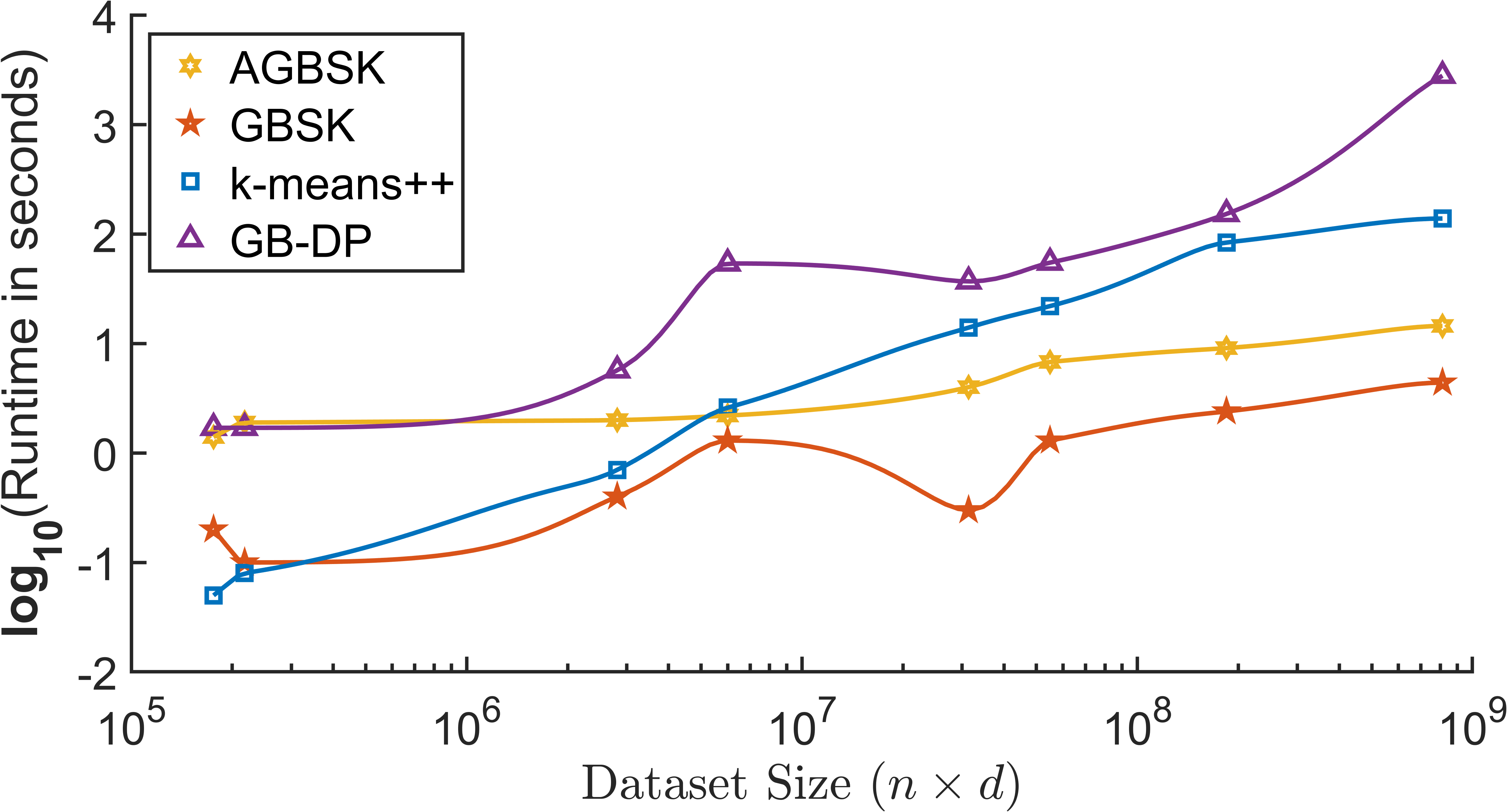}
    \caption{Logarithmic runtime comparison across various datasets. Refer to Table~\ref{tab:compare_GBSK_large} for details. }
    \label{fig:time_comp}
\end{figure}

DPeak and FSDPC require $d_c$ (cutoff distance) for local density estimation. 
GDPC uses $d_c$ and a screening ratio for grid-based computation. 
CDPC employs $d_c$ and a screening ratio to partition data into circular regions. 
FHC-LDP adjusts the kNN size based on dataset size. 
SDPC recommends the sampling scale $p$, nearest-neighbor scale $K$, and search expansion factor $K'$.

As shown in Table~\ref{tab: clustering quality}, the performance of GBSK is very competitive and better than most other algorithms, falling just slightly short of FHC-LDP on the \textsc{Segmentation} dataset. 

\begin{figure}[th]
    \centering
    \includegraphics[width=1\linewidth]{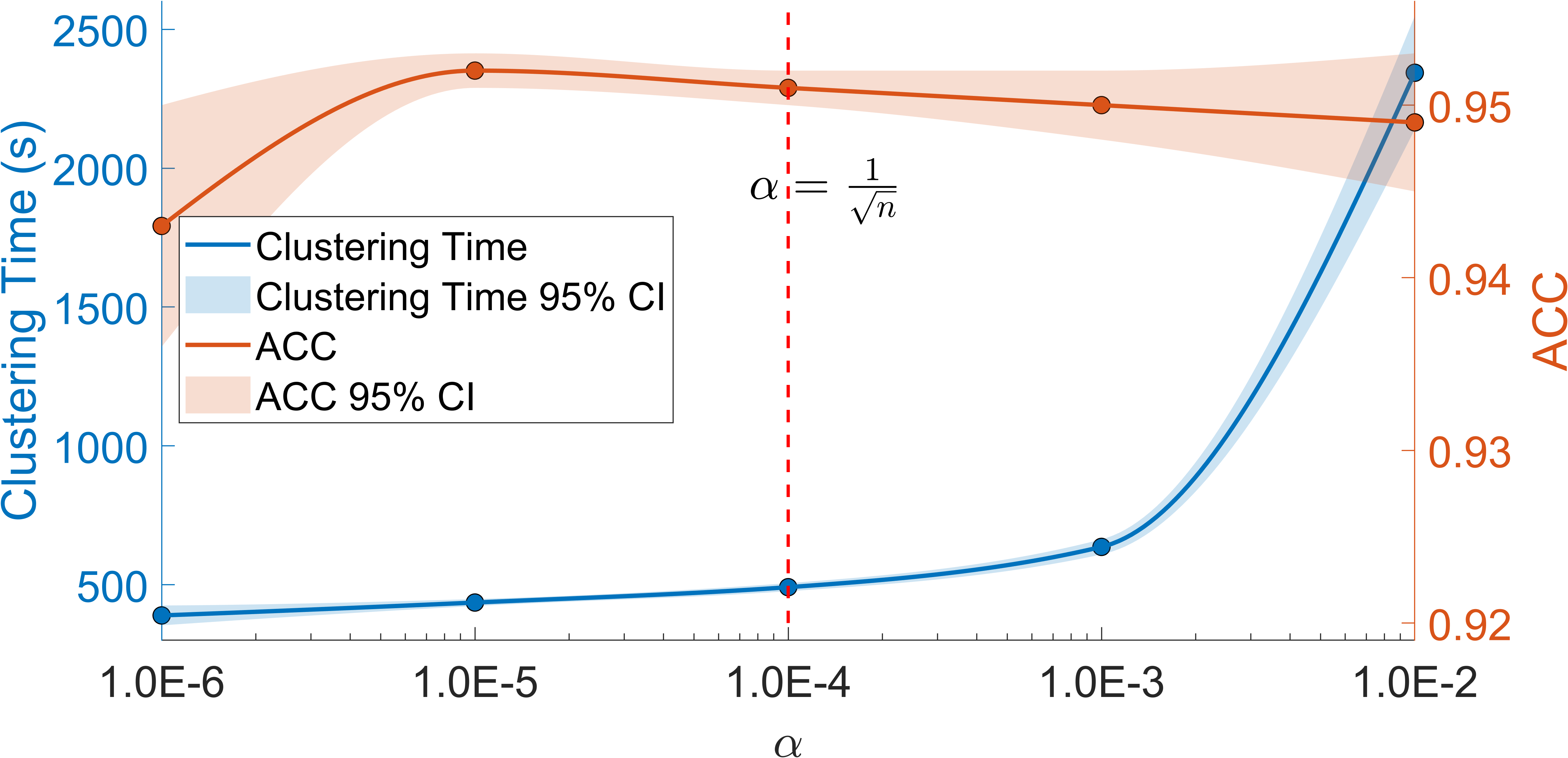}
    \caption{\textbf{Impact of $\alpha$} on runtime (s) and accuracy under $[30,*,10k,k]$ setting for GBSK on \textsc{AGC100M}, averaged over 50 runs with 95\% confidence intervals. The vertical dashed red line highlights $\alpha = \frac{1}{\sqrt{n}} = \num{1e-4}$. }
    \label{fig:AGC100M_alpha}
\end{figure}

\begin{figure}[th]
    \centering
    \includegraphics[width=1\linewidth]{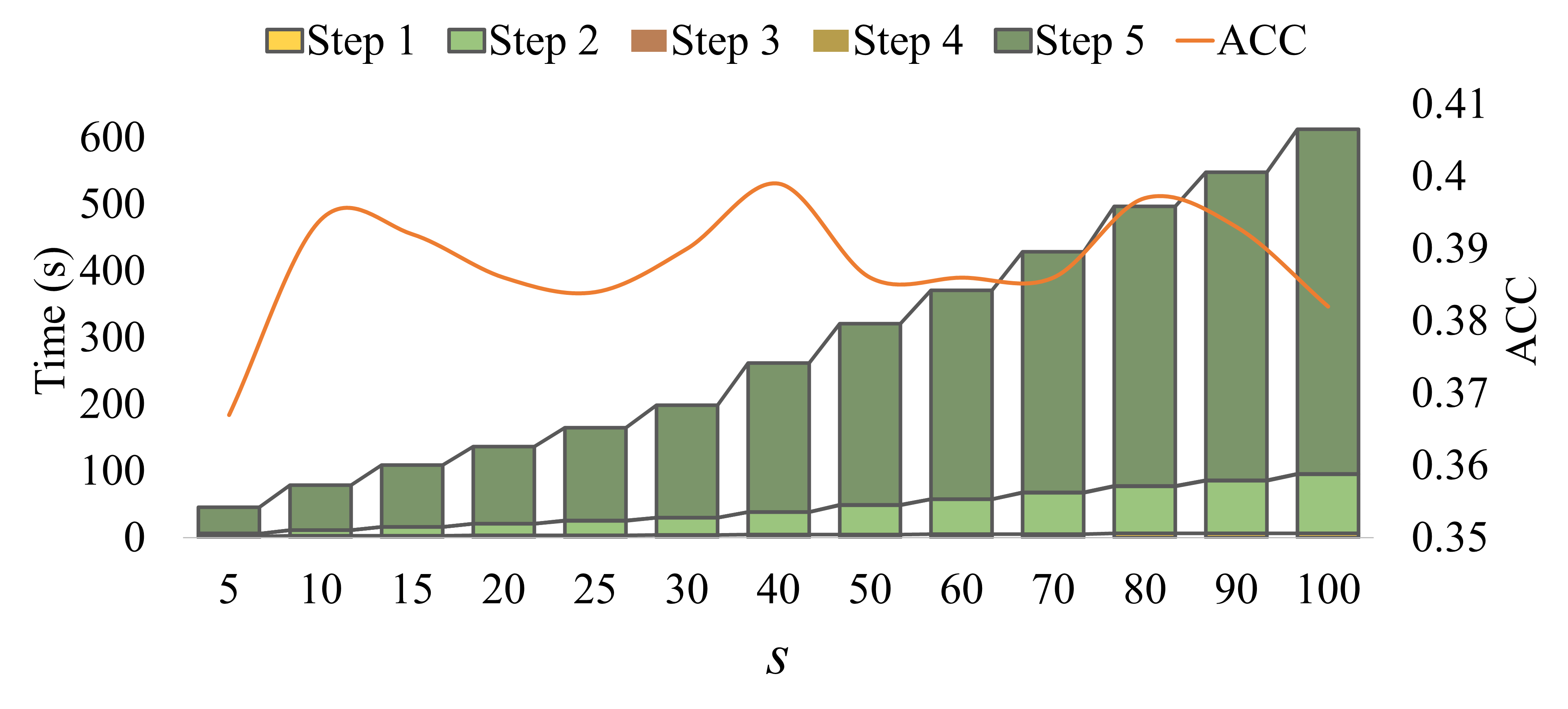}
    \caption{\textbf{Impact of $s$} on 
the time (in seconds) and accuracy for each step under $[*,\frac{1}{\sqrt{n}},10k,k]$ setting for GBSK on \textsc{MNIST8M}, averaged over 50 runs.}
    \label{fig:MNIST8M_s}
\end{figure}

\begin{figure}[!th]
    \centering
    \includegraphics[width=1\linewidth]{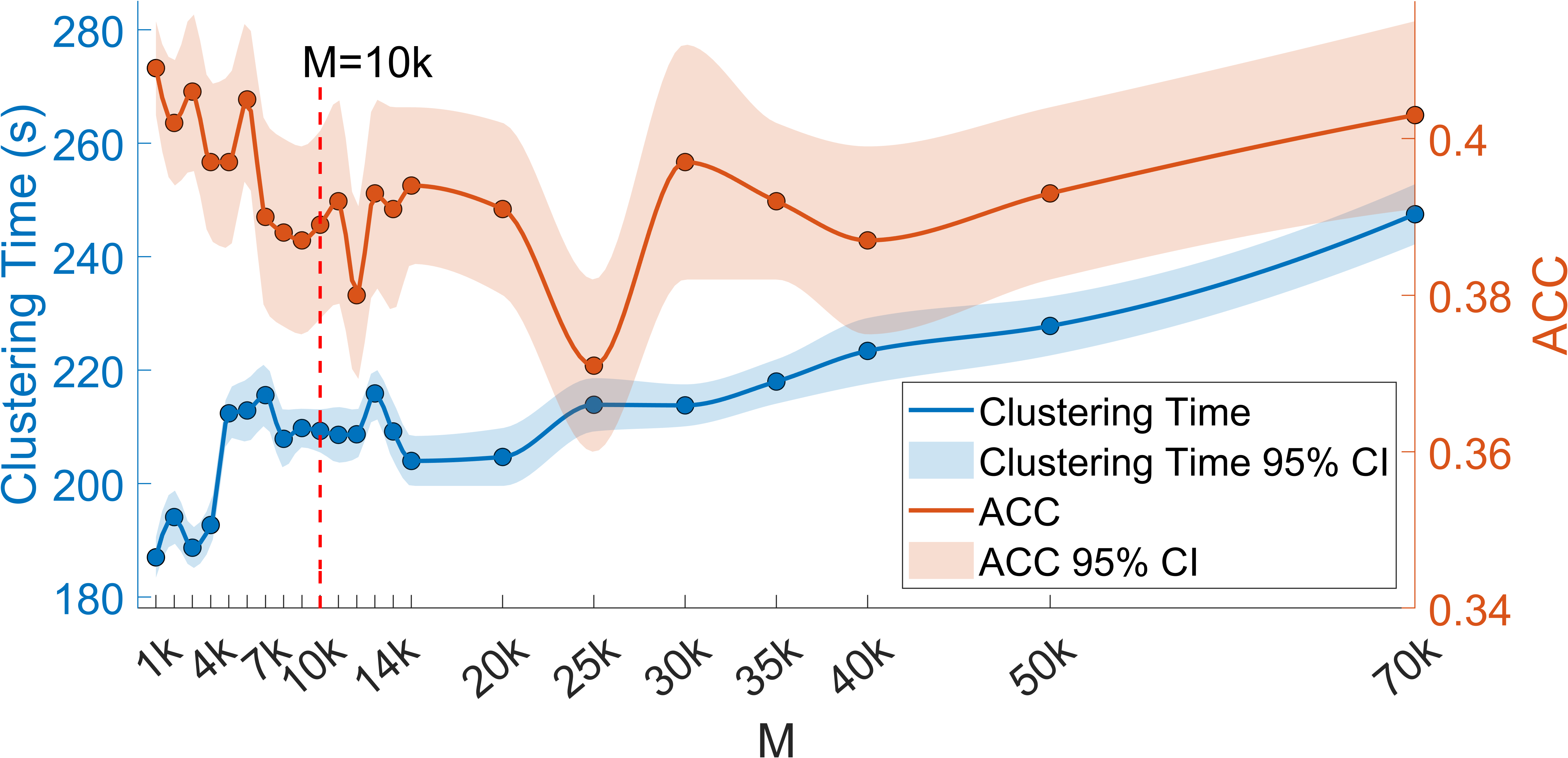}
    \caption{\textbf{Impact of $M$} on runtime (s) and accuracy under $[30,\frac{1}{\sqrt{n}},*,k]$ setting for GBSK on \textsc{MNIST8M}, averaged over 50 runs with 95\% confidence intervals. The vertical dashed red line highlights $M=10k=100$.}
    \label{fig:MNIST8M_M}
\end{figure}

\subsection{Comparisons on Large-scale Datasets}
In this subsection, we compare GBSK and AGBSK with FastDPeak, FHC-LDP, Fast LDP-MST, GB-DP, and k-means++ on several large-scale synthetic and real datasets. 

The parameter settings for each algorithm are well chosen based on their own recommendations of respective papers. AGBSK and k-means++ share the same parameter $k$, i.e., the number of clusters. 
FastDPeak, FHC-LDP, and Fast LDP-MST share two key parameters $K$ (the number of nearest neighbors for density estimation) and $C$ (the number of clusters). For simplicity, only $K$ of FastDPeak, FHC-LDP, and Fast LDP-MST is displayed in Table~\ref{tab:compare_GBSK_large} , while the number of clusters is kept consistent across each dataset.
GB-DP is parameter-free, with manual selection of $k$ cluster centers as the implicit parameter. The reported runtime excludes this interaction time. 

As shown in Table~\ref{tab:compare_GBSK_large}, GBSK achieves competitive runtime performance comparable to other algorithms while maintaining similar clustering quality on relatively small datasets, such as \textsc{Pendigits}, \textsc{DryBean}, and \textsc{MoCap}. However, as the dataset size increases, the advantages of GBSK become progressively more prominent. On very-large-scale datasets, including \textsc{CoverType}, \textsc{3M2D5}, \textsc{MNIST}, \textsc{CIFAR-10}, \textsc{MNIST8M}, and \textsc{AGC100M}, most competing methods either fail to execute or require prohibitively long runtimes.

Our comparative analysis reveals GB-DP, k-means++, GBSK, and AGBSK as the most computationally efficient algorithms. Fig.~\ref{fig:time_comp} (with data sourced from Table~\ref{tab:compare_GBSK_large}) demonstrates that their performance diverges significantly at scale. Notably, as dataset size exceeds $\num{6e6}$, GBSK and AGBSK outperform GB-DP and k-means++ while maintaining comparable accuracy. This underscores GBSK’s particular performance for very-large-scale, high-dimensional data. 

\subsection{Parameter Sensitivity Investigation in Large-scale Data}
We conduct a parameter grid search on three large-scale, high-dimensional datasets (\textsc{N-BaIoT}, \textsc{AGC100M}, \textsc{MNIST8M}) to assess the robustness and scalability of GBSK. The study evaluated how different parameter configurations impact performance, stability, and efficiency.
All results represent averages from 50 independent runs with 95\% confidence intervals. Complete experimental details are available in Appendices~B-D. 

\begin{enumerate}
    \item \textbf{$\alpha$}: The default value of $\alpha$ ($\frac{1}{\sqrt{n}}$) is able to yield near-optimal performance, as illustrated in Fig.~\ref{fig:AGC100M_alpha} and Table~\ref{tab:compare_GBSK_large}. This choice 
    balances computational efficiency with accurate data skeleton extraction. Deviating from this value leads to tradeoffs: larger $\alpha$ substantially increases runtime due to granular-ball generation overhead, while smaller $\alpha$ may, in certain cases, compromise structural fidelity. 
    
    \item \textbf{$s$}: As evidenced by Fig.~\ref{fig:MNIST8M_s} and Appendix~B Fig.~15, the runtime scales sub-linearly 
     w.r.t $s$, matching our theoretical complexity $O(s \times \sqrt{n} + W \times n)$. While excessively small $s$ (e.g., $s<10$) may lead to insufficient key balls ($W < k$), $s=30$ reliably achieves stable performance and representative granular-ball structures. The predominant computational cost is in Step 5, due to its highest complexity $O(W\times n)$, consistent with theoretical analysis in Section \ref{subsec:complxity analysis}. 
    
    \item \textbf{$M$}: Based on the experiments shown in Fig.~\ref{fig:MNIST8M_M} and Fig.~16 in Appendix~B,  we can see that Larger $M$ increases both ball-generation and label-assignment costs, and that $M=10k$ maintains stable accuracy while balancing computation time.
\end{enumerate}







\subsection{Ablation Study}
We evaluated the contribution of key components in GBSK through systematic ablation experiments across four datasets. All parameters of methods compared were selected in accordance with default configuration in Section~\ref{subsec:AGBSK_param}. Three variants are examined:

\begin{itemize}
    \item \textbf{GBSK without sampling}: 
    The variant (Algorithm~5 in Appendix~E) operates directly on the complete dataset without sampling, consequently bypassing representative ball selection, instead clustering using all generated granular-balls. 
    \item \textbf{GBSK without representative balls}: 
    This variant (Appendix~E Algorithm~6) performs sampling but skips representative balls and key balls, using original granular-balls for label assignment. 
    \item \textbf{GBSK without granular-ball}: 
    This variant (Appendix~E Algorithm~7) substitutes the granular-ball generation with density peak detection on two stages, maintaining GBSK's core architecture. 
\end{itemize}

\begin{figure}[t]
    \centering
        \begin{subfigure}{0.23\textwidth}
        \centering
        \includegraphics[width=\linewidth]{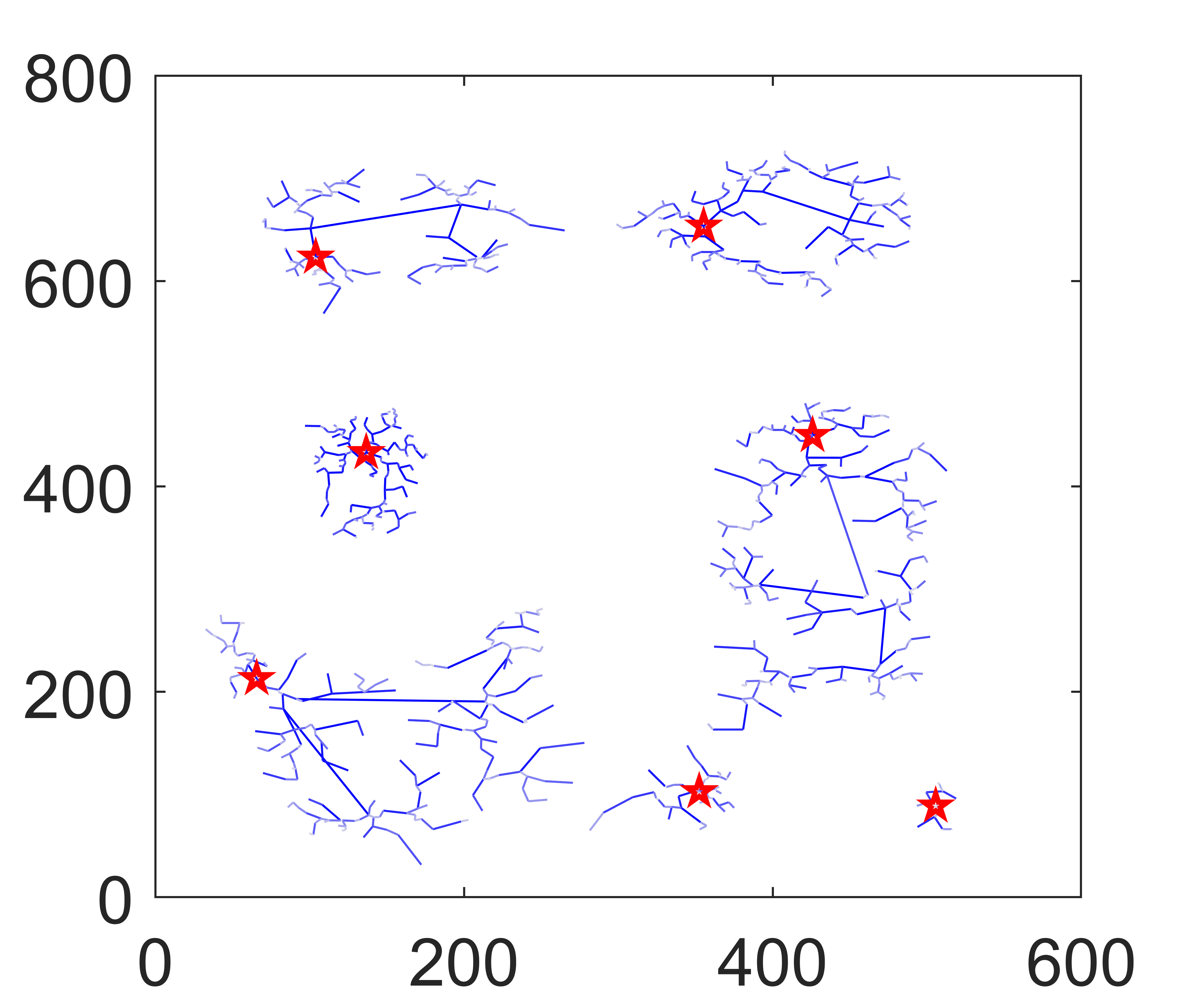}
        \caption{Original skeleton}
        \label{fig:Orig_Skel}
    \end{subfigure} 
    \begin{subfigure}{0.23\textwidth}
        \centering
        \includegraphics[width=\linewidth]{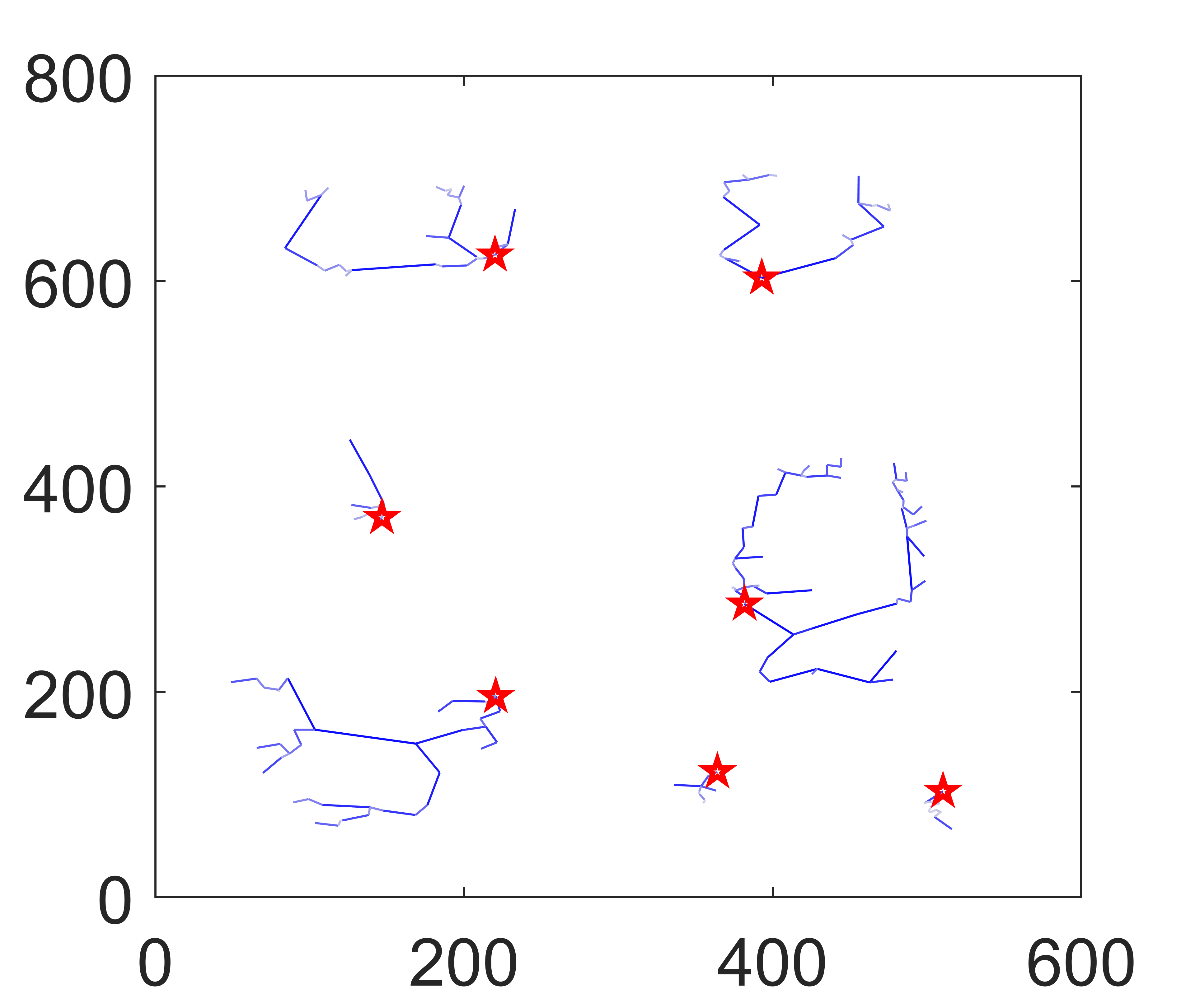}
        \caption{Standard skeleton}
        \label{fig:Std_Skel}
    \end{subfigure}
    
    \caption{\textbf{Effectiveness of representative balls on \textsc{SYN2}}. 
    (a) Skeleton drawn by \textit{AGBSK without representative balls} from granular-ball centers of all sample sets; 
    (b) Standard skeleton depicted by \textit{AGBSK} from aggregated representative centers, which is similar to (a) but much more succinct.}
    \label{fig: skeleton comparison}
\end{figure}

Table~\ref{tab:ablation}, Table \ref{tab:ablation_gb_number} and Fig. \ref{fig: skeleton comparison} reveal several findings:
\begin{itemize}
    \item \textit{Sampling importance}: The no-sampling variant fails on \textsc{AGC100M} due to memory constraints and shows substantially slower performance on larger datasets, demonstrating sampling's necessity for scalability. 
    \item \textit{Granular-ball advantage}: While the no-granular-ball variant shows faster execution on smaller datasets like \textsc{Pendigits} and \textsc{MNIST}, it exhibits consistently inferior clustering quality and poor scalability due to DPeak's quadratic complexity. This empirical evidence confirms that granular-balls provide both computational efficiency and competitive clustering quality.

    \item \textit{Representative ball efficiency}: As  Fig.~\ref{fig: skeleton comparison} shows, while both methods produce similar skeletons, the representative-ball version achieves this more compactly. Table~\ref{tab:ablation_gb_number} reveals that GBSK maintains a relatively stable number of representative balls regardless of dataset size. This stability reduces computational overhead in label assignment, explaining why the no-representative-balls variant runs significantly slower than GBSK.
    
\end{itemize}

\begin{table}
\setlength{\textfloatsep}{5pt plus 2pt minus 2pt}
\setlength{\floatsep}{5pt}
\setlength{\intextsep}{5pt}
    \centering
    \renewcommand{\arraystretch}{1.2}
    \setlength{\tabcolsep}{2.65pt}
    \caption{Ablation Comparison}
    \label{tab:ablation}
    \begin{threeparttable}
    \begin{tabular}{|c|c|c|c|c|c|}
    \hline
    \multirow{3}{*}{\rotatebox{90}{Dataset}} & \multirow{3}{*}{Metric} & \multicolumn{4}{c|}{Methods} \\
    \cline{3-6}
    & & \multirow{2}{*}{GBSK} & \multirow{2}{*}{\makecell{GBSK w/o \\sampling}} & \multirow{2}{*}{\makecell{GBSK w/o \\representative balls}} & \multirow{2}{*}{\makecell{GBSK w/o \\granular-balls}} \\
    &&&&&\\
    \hline
    \multirow{4}{*}{\rotatebox{90}{\textsc{Pendigits}}} 
    & ACC & \textbf{0.796} & 0.727 & 0.754 & 0.763 \\
    \cline{2-6}
    & ARI & \textbf{0.644} & 0.602 & 0.627 & 0.619 \\
    \cline{2-6}
    & AMI & 0.732 & \textbf{0.739} & 0.726 & 0.711 \\
    \cline{2-6}
    & Time & 1.41 s & 0.46 s & 1.66 s & \textbf{0.04} s \\
    \hline
    \multirow{4}{*}{\rotatebox{90}{\textsc{MNIST}}} 
    & ACC & \textbf{0.557} & 0.550 & 0.551 & 0.448 \\
    \cline{2-6}
    & ARI & \textbf{0.405} & 0.362 & 0.362 & 0.234 \\
    \cline{2-6}
    & AMI & \textbf{0.544} & 0.496 & 0.489 & 0.351 \\
    \cline{2-6}
    & Time & 6.82 s & 30.10 s & 22.32 s & \textbf{2.51} s \\
    \hline
    \multirow{4}{*}{\rotatebox{90}{\textsc{N-BaIot}}} 
    & ACC & \multicolumn{4}{c|}{} \\
    \cline{2-2}
    & ARI & \multicolumn{4}{c|}{without ground truth labels} \\
    \cline{2-2}
    & AMI & \multicolumn{4}{c|}{} \\
    \cline{2-6}
    & Time & \textbf{14.46} s & 362.40 s & 397.79 s & 54.54 s \\
    \hline
    \multirow{4}{*}{\rotatebox{90}{\textsc{MNIST8M}}} 
    & ACC & \textbf{0.521} & 0.437 & 0.387 & 0.338 \\
    \cline{2-6}
    & ARI & \textbf{0.308} & 0.258 & 0.228 & 0.141 \\
    \cline{2-6}
    & AMI & \textbf{0.426} & 0.415 & 0.406 & 0.212 \\
    \cline{2-6}
    & Time & \textbf{184.13} s & 8733.12 s & 5856.43 s & 538.62 s \\
    
    \hline
    \multirow{4}{*}{\rotatebox{90}{\textsc{AGC100M}}} & ACC & 0.951 & \multirow{4}{*}{out of memory} & \textbf{0.962} & 0.950 \\
    \cline{2-3}\cline{5-6}
    & ARI & 0.946 &  & \textbf{0.967} & 0.930 \\
    \cline{2-3}\cline{5-6}
    & AMI & 0.986 &  & \textbf{0.989} & 0.943 \\
    \cline{2-3}\cline{5-6}
    & Time & \textbf{491.23} s &  & 27877.79 s & 5535.22 s \\
    \hline
    \end{tabular}

    \end{threeparttable}
\end{table}

\begin{table}
\setlength{\textfloatsep}{5pt plus 2pt minus 2pt}
\setlength{\floatsep}{5pt}
\setlength{\intextsep}{5pt}
    \centering
    \renewcommand{\arraystretch}{1.2}
    \setlength{\tabcolsep}{2.65pt}
    \caption{Comparison on number of key components.}
    \label{tab:ablation_gb_number}
    \begin{threeparttable}
    \begin{tabular}{|c|c|c|c|c|}
    \hline
    \multirow{5}{*}{Dataset} & \multicolumn{4}{c|}{Methods} \\
    \cline{2-5}
    & \multirow{2}{*}{GBSK} & \multirow{2}{*}{\makecell{GBSK w/o\\sampling}} & \multirow{2}{*}{\makecell{GBSK w/o\\representative balls}} & \multirow{2}{*}{\makecell{GBSK w/o\\granular-balls}} \\
    &&&&\\
    \cline{2-5}
    & \makecell{\#repBalls} & \makecell{\#granular-balls} & \makecell{\#granular-balls} & \makecell{\#peaks}\\
    \hline
    \textsc{Pendigits} & 282 & 254 & 919 & 287 \\
    \hline
    \textsc{MNIST} & 300 & 149 & 2091 & 296 \\
    \hline
    \textsc{MNIST8M} & 300 & 135 & 3777 & 300 \\
    \hline
    \textsc{N-BaIot} & 250 & 148 & 1985 & 267 \\
    \hline
    \textsc{AGC100M} & 254 & \makecell{out of \\memory} & 8531 & 509 \\
    \hline
    \end{tabular}
    \begin{tablenotes}
        \item ``\#repBalls'': number of representative balls from all sample sets. 
        \item ``\#peaks'': number of $peaks$ found from all sample sets, which is similar to representative balls (see Algorithm~7). 
    \end{tablenotes}
    \end{threeparttable}
\end{table}

\section{CONCLUSION}\label{sec:conclusion}

In this work, we introduce a novel super fast clustering algorithm GBSK, as well as its simplified version AGBSK, designed for large-scale datasets and capable of running efficiently on standard computing hardware. The superiority stems from its innovative integration of granular-ball computing and multi-sampling strategy. By multi-sampling the dataset and constructing multi-grained granular-balls, they progressively uncover a statistical ``skeleton''  of the original data, which could maintain the essential abstraction of the original data, while dramatically reduce computational cost for clustering. 

GBSK and AGBSK achieve superior performance across three key aspects: effectiveness, efficiency, and scalability. First, they maintain competitive clustering accuracy compared to traditional methods while operating on compressed data representations. Second, the algorithms exhibit remarkable computational efficiency, processing massive datasets with sub-linear time complexity, significantly outperforming state-of-the-art methods in terms of speed. Most importantly, both methods demonstrate excellent scalability - their time and space complexity scale linearly or sub-linearly with dataset size, as verified through controlled experiments on progressively larger datasets. This combination of accuracy and efficiency makes our approach particularly suitable for real-world applications involving large-scale data processing.


\section*{Acknowledgments}
This work was supported in part by the National Natural Science Foundation of China under Grant No.61673186, 62221005 and 62376045.

\vfill
\bibliographystyle{IEEEtran}

\end{document}